\documentclass[letterpaper]{article} 
\usepackage[draft]{aaai2026}  
\usepackage{times}  
\usepackage{helvet}  
\usepackage{courier}  
\usepackage[hyphens]{url}  
\usepackage{graphicx} 
\usepackage{natbib}  
\usepackage{caption} 
\usepackage{algorithm}
\usepackage{algorithmic}

\usepackage{amsfonts}
\usepackage{amsmath}
\usepackage{booktabs} 
\usepackage{amssymb}
\usepackage{graphicx}
\usepackage{subcaption}
\usepackage{pifont}
\newcommand{\cmark}{\ding{51}}  
\newcommand{\xmark}{\ding{55}}  
\usepackage[utf8]{inputenc} 
\usepackage{newfloat}
\usepackage{listings}
\DeclareCaptionStyle{ruled}{labelfont=normalfont,labelsep=colon,strut=off} 
\floatstyle{ruled}
\newfloat{listing}{tb}{lst}{}
\floatname{listing}{Listing}
\title{IBEX: Information-Bottleneck-EXplored Coarse-to-Fine Molecular Generation under Limited Data}

\author{
{\fontsize{11pt}{11pt}\selectfont
  Dong Xu\textsuperscript{\rm 1,2}\equalcontrib,~
  Zhangfan Yang\textsuperscript{\rm 3}\equalcontrib,~
  Jenna Xinyi Yao\textsuperscript{\rm 4},~
  Shuangbao Song\textsuperscript{\rm 5},~
  Zexuan Zhu\textsuperscript{\rm 1,2},~
  Junkai Ji\textsuperscript{\rm 1,2}\thanks{Corresponding author.}
}
}
\affiliations{
  \textsuperscript{\rm 1}School of Artificial Intelligence, Shenzhen University, Shenzhen, China\\
  \textsuperscript{\rm 2}National Engineering Laboratory for Big Data System Computing Technology, Shenzhen University, Shenzhen, China\\
  \textsuperscript{\rm 3}School of Computer Science, University of Nottingham Ningbo, Ningbo, China\\
  \textsuperscript{\rm 4}Biology Department, University of California San Diego, California, USA\\
  \textsuperscript{\rm 5}School of Computer Science and Artificial Intelligence, Changzhou University, Changzhou, China\\[2pt]
  2400671001@mails.szu.edu.cn,\quad
  yzfshuaige@gmail.com,\quad
  jijunkai@szu.edu.cn
}

\usepackage{bibentry}

\begin{document}

\maketitle

\begin{abstract}
Three-dimensional generative models increasingly drive structure-based drug discovery, yet it remains constrained by the scarce publicly available protein–ligand complexes. Under such data scarcity, almost all existing pipelines struggle to learn transferable geometric priors and consequently overfit to training-set biases. As such, we present IBEX, an Information-Bottleneck-EXplored coarse-to-fine pipeline to tackle the chronic shortage of protein–ligand complex data in structure-based drug design. Specifically, we use PAC-Bayesian information-bottleneck theory to quantify the information density of each sample. This analysis reveals how different masking strategies affect generalization and indicates that, compared with conventional de novo generation, the constrained Scaffold Hopping task endows the model with greater effective capacity and improved transfer performance. IBEX retains the original TargetDiff architecture and hyperparameters for training to generate molecules compatible with the binding pocket; it then applies an L-BFGS optimization step to finely refine each conformation by optimizing five physics-based terms and adjusting six translational and rotational degrees of freedom in under one second. With only these modifications, IBEX raises the zero-shot docking success rate on CBGBench CrossDocked2020-based from 53\% to 64\%, improves the mean Vina score from \(-7.41\) kcal mol\(^{-1}\) to \(-8.07\) kcal mol\(^{-1}\), and achieves the best median Vina energy in 57 of 100 pockets versus 3 for the original TargetDiff. IBEX also increases the QED by 25\%, achieves state-of-the-art validity and diversity, and markedly reduces extrapolation error.
\end{abstract}

\section{Introduction}

\begin{figure}[!h]
\setlength{\intextsep}{0pt}  
\setlength{\abovecaptionskip}{4pt} 
\centering
\includegraphics[width=0.45\textwidth]{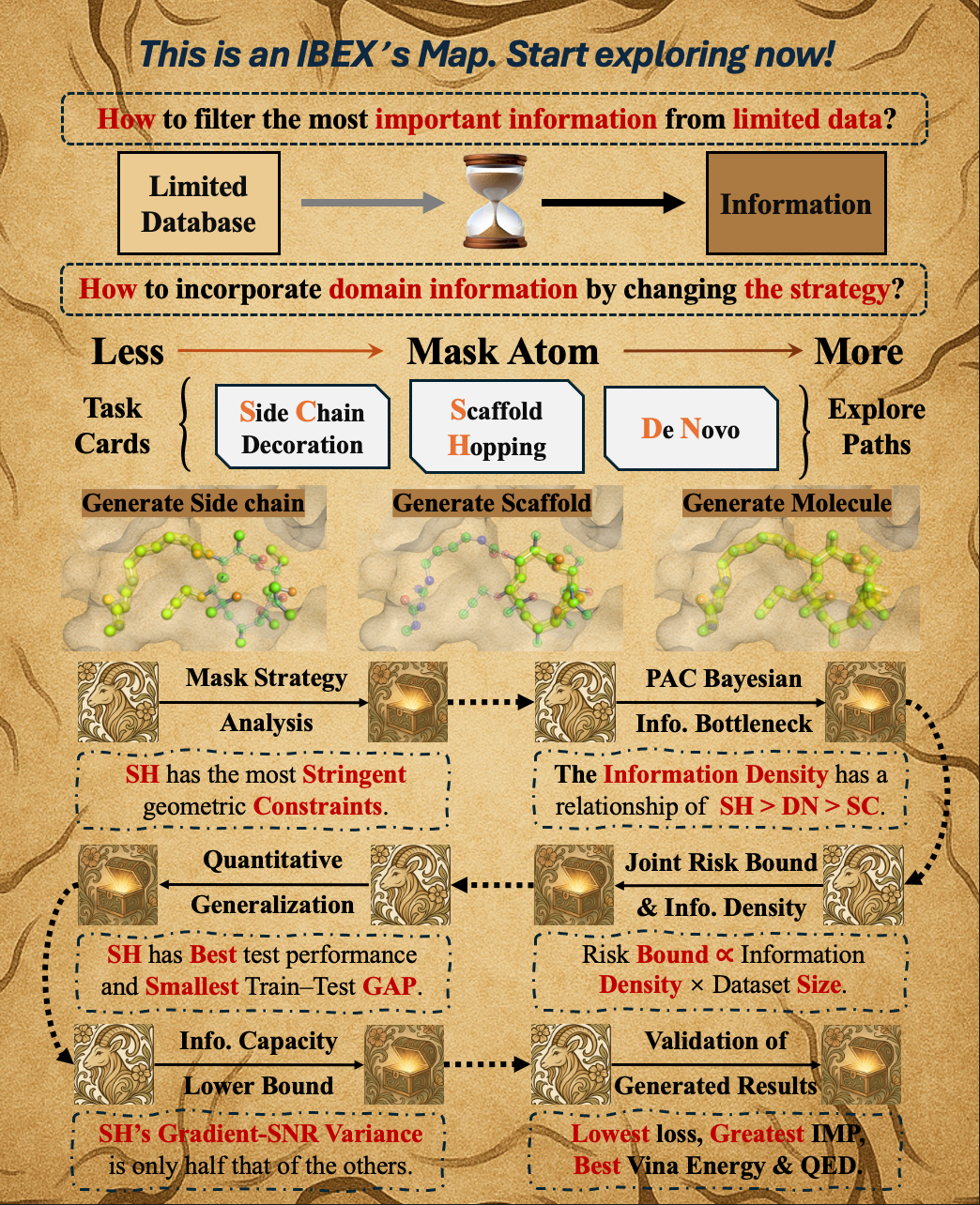}
\caption{Conceptual overview of the IBEX pipeline. Starting from a limited protein–ligand database, IBEX applies three masking strategies (SC, SH and DN) to maximally filter mutual information under a PAC-Bayesian bottleneck. Analysis reveals an information-density ordering $\rho_{\rm SH}>\rho_{\rm DN}>\rho_{\rm SC}$, tightening the test‐risk bound $\mathcal{R}\propto\rho\!\times\!|D|$. The SH task, with the most stringent geometric constraints, exhibits the lowest gradient‐SNR variance, the earliest capacity compression phase, and the smallest train–test gap.}

\label{fig:overall}
\end{figure}

Small-molecule discovery is leaving the classical ``virtual screening and lead optimization'' and moving toward target-aware design driven by three-dimensional generative models~\cite{CASDD}. Drug chemistry, however, faces a severe data bottleneck: fewer than  \(2\times10^{5}\) experimentally validated protein–ligand complexes are public~\cite{pdbbind}, while vision~\cite{dalle3} and language models~\cite{bert, gpt3} rely on corpora that are three orders of magnitude larger. 
The high cost of acquiring new complexes forces us to mine as much information as possible from each limited example. 
Protein–ligand co-folding models primarily memorize training-set biases rather than learning genuine binding preferences~\cite{P-L_co-folding, co-folding_docking}.
They remain insensitive to complete pocket-residue mutagenesis or side-chain polarity inversion~\cite{co-folding-_inter}.
AlphaFold3, being ligand-agnostic and trained solely on backbone conformations, likewise cannot overcome these biases and thus fails to accurately predict authentic protein–ligand binding modes~\cite{AlphaFold3}.

Most current 3D diffusion models follow a de novo protocol. They mask the entire ligand and regenerate it inside the protein pocket. 
Each sample therefore gives only a coarse prior— indicating possible atomic placements— and rarely conveys the core geometric rules that link pocket shape to molecular scaffold. End-to-End schemes that merge generation and docking inherit this weakness. 
Gradient signals become diluted, physical interpretability drops, and binding poses are often sub-optimal. 
Even the best standalone docking tools still show limited placement accuracy and strong reliance on known motifs.

We introduce \textbf{IBEX} (\textbf{I}nformation-\textbf{B}ottleneck-\textbf{EX}plored), a two-stage framework that separates information-rich generation from physics-guided refinement. 
\textbf{High-information.} IBEX keeps key functional groups fixed and lets the model rebuild the molecular core. 
Anchoring these groups shrinks the search space and increases mutual information. 
The model thus learns richer priors from fewer samples and can generalize to de novo without extra tuning.
\textbf{Coarse-to-fine.} After sampling candidate molecules inside the pocket, we treat the ligand as a rigid body and run a limited-memory BFGS search. 
The optimiser jointly minimises van-der-Waals attraction, steric repulsion, and hydrogen-bond energy. 
Decoupling this step keeps the physical objective clear and avoids the gradient dilution seen in end-to-end schemes.

Information-theoretic analysis shows that scaffold-hopping data provide IBEX with higher information density, a tighter PAC-Bayes information-bottleneck bound, and an effectively larger sample. 
Experiments on geometry settings, model capacity, and diffusion scores demonstrate robust zero-shot transfer without parameter updates.

Our main contributions are:
\begin{itemize}
  \item The first explicit risk bound based on physicochemical information density, filling a gap in quantifying model generalization under extreme data scarcity.
  \item The first application of gradient SNR ratio analysis in 3D molecular generation, revealing a clear scaling relationship among risk, model capacity, and information.
  \item The first demonstration of zero‐shot transfer from limited scaffold‐hopping training to de novo molecular generation, attaining state‐of‐the‐art performance under extreme data scarcity without additional fine‐tuning.
  \item A new coarse‐to‐fine generation–physical refinement paradigm for future structure‐based drug design, centered on information bottleneck theory and efficiently coupling information‐theoretic principles with physics‐based optimization.
\end{itemize}

\section{Related Work}
\label{sec:related-work}

Molecular generative modelling has advanced rapidly in recent years. Broadly, contemporary approaches fall into two categories: \emph{target‐agnostic} models that explore chemical space without reference to specific proteins, and \emph{target‐aware} models that design ligands in the presence of an explicit binding site or pocket.

\textbf{Target‐Free Molecular Generation.}
Target‐free generators are judged mainly by chemical validity, diversity, and drug‐likeness. 
SE(3)-equivariant diffusion generates either graphs or full 3D coordinates without protein by reversing a noise process \cite{xu2022geodiff, hoogeboom2022equivariant, morehead2024geometry}. 
GraphAF combines flows with autoregression for goal-directed sampling \cite{graphAF}, while GraphDF uses discrete flows to better cover the combinatorial space \cite{graphdf}. 
Scaffold-aware variants narrow the search by fixing or first generating a core: Lim~et al. retain a user-specified scaffold during atom-wise growth \cite{lim2020scaffold}; Sc2Mol divides the task into VAE scaffold discovery followed by Transformer decoration \cite{sc2mol}; and fragment-hierarchical methods such as MolPAL and Junction-Tree VAE build coarse fragments or trees before atomic refinement \cite{molpal,jtvae}. 

\textbf{Target‐Aware Molecular Generation.}
Structure-based drug-design models condition generation on pocket geometry. 
DiffSBDD pioneered pocket-aware denoising diffusion \cite{diffsbdd}, TargetDiff added an affinity term to bias toward tight binders \cite{targetdiff}, and DiffBP removed sequential bias via whole-molecule denoising \cite{diffbp}; hierarchical extensions D3FG and DecompDiff diffuse functional groups or scaffold-arm decompositions for improved geometry and synthesizability \cite{d3fg,decompdiff}. 
Autoregressive pocket-conditioned approaches place atoms step-wise: Pocket2Mol uses an E(3)-equivariant GNN \cite{pocket2mol}, GraphBP deploys a local flow model \cite{graphbp}, ResGen incorporates residue-level encoding \cite{resgen}, and TamGen employs a GPT-style chemical language model for rapid SMILES generation \cite{tamgen}. 
Fragment-centric variants further constrain chemistry while maintaining flexibility: FLAG sequentially inserts predefined fragments into the pocket \cite{flag}, MolCRAFT performs continuous 3D optimization before collapsing to a discrete ligand \cite{molcraft}, and linker methods such as Delinker and FragGrow extend anchored pharmacophores \cite{delinker,fraggrow}. 

\section{Methods}
\label{Sec:method}

\begin{figure}[t]
\centering
\includegraphics[width=0.48\textwidth]{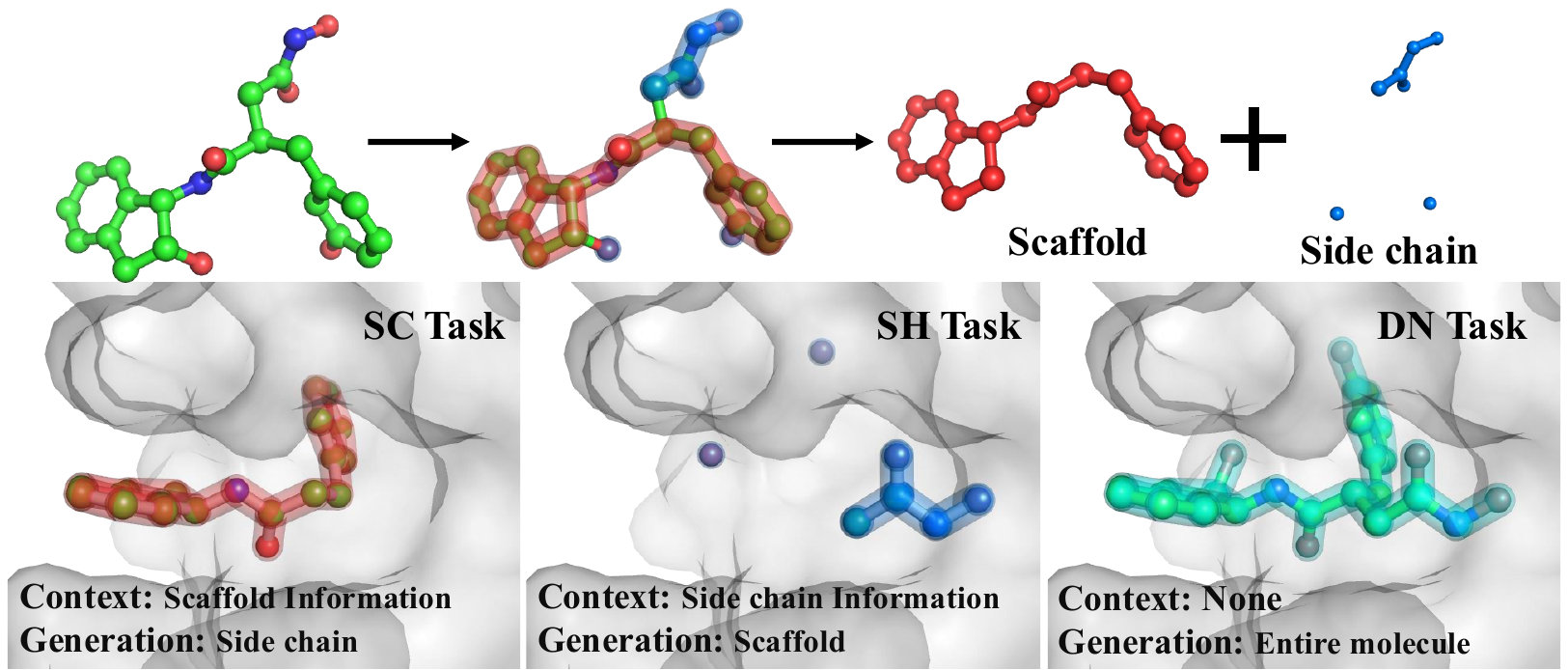}
\caption{\textbf{Workflow of the Bemis–Murcko decomposition.} 
Starting from the full ligand structure, the algorithm (i) identifies all ring systems and the linkers that connect them, (ii) removes peripheral side chains that are not part of a ring or linker, and (iii) collapses any fused ring junctions to generate the unique Bemis–Murcko scaffold.}
\label{fig:bemis}
\end{figure}

\textbf{Notation and preliminaries.}
To formalize the three generative scenarios considered in this work—scaffold hopping (\textsc{SH}), side‑chain decoration (\textsc{SC}), and de‑novo generation (\textsc{DN})—we first decompose every ligand \(M\) bound to a pocket \(P\) via a three‑step Bemis–Murcko reduction (Figure~\ref{fig:bemis} Top). 
Ring systems and their connecting linkers are enumerated, peripheral atoms are excised, and fused junctions are merged, leaving a canonical scaffold \(\mathcal{S}\) (yellow) and a complementary side‑chain set \(\mathcal{C}\) (blue). 
Conditioning on these fragments yields the task indicator 
\(\mathcal{T}\in\{\textsc{SH},\textsc{DN},\textsc{SC}\}\):
\textsc{SH} receives \(\mathcal{C}\) and proposes alternative scaffolds, 
\textsc{SC} fixes \(\mathcal{S}\) and generates diverse \(\mathcal{C}'\), 
whereas \textsc{DN} samples an entire ligand without prior structural constraints. 

Each task has an associated dataset 
\(D_{\mathcal{T}}=\{(P_i,M_i)\}_{i=1}^{N_{\mathcal{T}}}\) 
with empirical joint density \(\hat q_{\mathcal{T}}(P,M)\); 
\(N_{\mathcal{T}}=\lvert D_{\mathcal{T}}\rvert\) denotes its size. 
A shared SE(3)‑equivariant diffusion generator \(G_{\theta}\) and a docking/refinement module \(D_{\phi}\) act on all tasks, and their performance is evaluated via empirical and population risks \(\widehat{\mathcal{R}}\) and \(\mathcal{R}\). 
Information‑theoretic quantities such as differential entropy \(H(\cdot)\) and mutual information \(I(\cdot\,;\cdot)\) are reported in natural‑log units. 

A ligand and pocket are represented by their atom sets as follows~\cite{targetdiff}:
\begin{equation}
    \mathcal{S}_M=\bigl\{(\mathbf{x}_M^{(i)},\mathbf{v}_M^{(i)},c_M^{(i)})\bigr\}_{i=1}^{N_M},
\end{equation}
\begin{equation}
\mathcal{S}_P=\bigl\{(\mathbf{x}_P^{(j)},\mathbf{v}_P^{(j)},b_P^{(j)},r_P^{(j)})\bigr\}_{j=1}^{N_P},
\end{equation}
where \(N_M\) and \(N_P\) are the ligand‑atom and pocket‑atom counts. 
Each atom carries Cartesian coordinates \(\mathbf{x}\in\mathbb{R}^{3}\) and an element‑type one‑hot vector \(\mathbf{v}\in\mathbb{R}^{K}\) from a vocabulary of size \(K\). 
For ligand atoms, the binary flag \(c\) marks whether the atom is fixed by the task context (\(c=1\)); for pocket atoms, \(b\) indicates backbone membership, and \(r\in\mathbb{R}^{K'}\) is a one‑hot vector over \(K'\) amino‑acid residues. 
Stacking these features row‑wise yields the matrices
\(\mathbf{m}=[\mathbf{X}_M,\mathbf{V}_M,\mathbf{c}_M]\in\mathbb{R}^{N_M\times(3+K+1)}\) 
and 
\(\mathbf{p}=[\mathbf{X}_P,\mathbf{V}_P,\mathbf{b}_P,\mathbf{r}_P]\in\mathbb{R}^{N_P\times(3+K+1+K')}\),
which serve as the inputs to \(G_{\theta}\) and \(D_{\phi}\).

\textbf{3D Pocket-aware Diffusion as a generator.}
The generator keeps TargetDiff backbone and introduces two forward noise channels—Gaussian for coordinates and categorical for atom types. 
For each pair \((P,M)\) a spatial mask \(M_{\mathrm{tgt}}\subseteq\{1,\dots,N_M\}\) is sampled; indices in \(M_{\mathrm{tgt}}\) are regenerated, the rest form the context \(M_{\mathrm{ctx}}\).
Let
\(\mathbf{x}_0=M_{\mathrm{tgt}}^{\text{x}}\in\mathbb{R}^{3\times\lvert M_{\mathrm{tgt}}\rvert}\)
and
\(\mathbf{v}_0=M_{\mathrm{tgt}}^{\text{v}}\in\mathbb{R}^{K\times\lvert M_{\mathrm{tgt}}\rvert}\)
be their clean coordinates and types.
The forward noising at step \(t\) is
\begin{align}
q_t(\mathbf{x}_t\mid\mathbf{x}_0) &=
\mathcal{N}\!\bigl(\sqrt{\bar\alpha_t}\,\mathbf{x}_0,\,
(1-\bar\alpha_t)\mathbf{I}\bigr), \label{eq:coord_noise}\\[4pt]
q_t(\mathbf{v}_t\mid\mathbf{v}_0) &=
\mathcal{C}\!\bigl(\bar\alpha_t\,\mathbf{v}_0 + (1-\bar\alpha_t)/K\bigr), 
\label{eq:type_noise}
\end{align}
where \(\bar\alpha_t=\prod_{s=1}^{t}\alpha_s\) is the cumulative variance schedule and \(\mathcal{C}(\cdot)\) denotes a categorical distribution over the \(K\) atom types~\cite{targetdiff, diffbp}. 
Protein coordinates are weakly perturbed for regularisation~\cite{Dockformer}:
\begin{equation}
\tilde{\mathbf{x}}_P=\mathbf{x}_P+\boldsymbol{\varepsilon}, \quad
\boldsymbol{\varepsilon}\sim\mathcal{N}\!\bigl(\mathbf{0},0.1^{2}\mathbf{I}\bigr). \label{eq:prot_noise}
\end{equation}

The reverse process uses two heads:
\(s_\theta^{\text{x}}(P,\mathbf{x}_t,t)\) predicts the coordinate score, and
\(s_\theta^{\text{v}}(P,\mathbf{v}_t,t)\) predicts type logits.
The total loss is the sum of coordinate and type objectives:
\begin{align}
\mathcal{L}_{\text{x}}(\theta)= 
\mathbb{E}_{t,(P,M)} \Bigl[\lambda_t \bigl\lVert s_\theta^{\text{x}}- \nabla_{\mathbf{x}_t}\log q_t(\mathbf{x}_t\mid\mathbf{x}_0)\bigr\rVert_{2}^{2}\Bigr],\\
\mathcal{L}_{\text{v}}(\theta)= 
\mathbb{E}_{t,(P,M)}\Bigl[\gamma_t \operatorname{CrossEntropy}\bigl(s_\theta^{\text{v}},\mathbf{v}_0\bigr)\Bigr], \label{eq:diff_loss}
\end{align}
with \(\lambda_t=\sigma_t^{2}/\alpha_t^{2}\), \(\sigma_t^{2}=1-\alpha_t\), and \(\gamma_t\) mirroring the type-noise variance.
At inference, ancestral sampling yields a coarse pose \(M_0 = G_\theta(P)\) whose heavy atoms fall inside a \(10\,\text{\AA}\) sphere centred on the pocket.

\textbf{Physics-guided Position Refinement (PR).}
Given a coarse ligand proposal \(M_0\) from the generator, we refine its placement in the pocket using a lightweight, gradient-based search over rigid-body degrees of freedom.
During this search, the ligand is treated as a rigid object; only global rotations and translations are updated, whereas internal covalent geometry, atom types, and formal charges remain fixed.
This design isolates pose quality from generative uncertainty and concentrates limited gradient signal on a six-dimensional space (3 translation, 3 rotation). 

The refinement minimizes a differentiable surrogate of the binding free energy constructed from five short-range physical contact terms commonly used in empirical coupling potentials: two Gaussians, hard-sphere penalty, hydrophobic, hydrogen-bond. For details, please refer to Appendix D. The total energy is a weighted sum
\begin{equation}
E_{\text{phys}}(P,M) \;=\;
W^{T}(E_{\text{g1}} + E_{\text{g2}} + E_{\text{rep}} + E_{\text{hyd}} + E_{\text{hd}}),
\label{eq:phys_energy}
\end{equation}
where lower values indicate better predicted affinity.
Weights \(W^{T}\) are held fixed across all experiments and were set once on a small calibration panel~\cite{Vina}.
Let \(\mathbf{X}\in\mathbb{R}^{3\times N_M}\) be ligand coordinates in the generator frame and \(\mathbf{R}\in\mathrm{SO}(3)\), \(\mathbf{t}\in\mathbb{R}^3\) be the current rigid transform~\cite{carsidock}.
The refined pose is
\begin{equation}
\mathbf{X}' = \mathbf{R}\,\mathbf{X} + \mathbf{t}\mathbf{1}^\top.
\end{equation}
We parameterise \(\mathbf{R}\) by an axis–angle 3-vector (Rodrigues) and optimise the 6-vector 
\(\mathbf{u} = (\delta_x,\delta_y,\delta_z,\omega_x,\omega_y,\omega_z)\).
Small updates compose via exponential maps; for clarity, we write this as 
\(\mathbf{R}_{k+1} = \exp(\boldsymbol{\omega}_k^\times)\mathbf{R}_k\) with \(\boldsymbol{\omega}_k=(\omega_x,\omega_y,\omega_z)\) and skew operator \((\cdot)^\times\).
The physics score is computed by an external energy evaluator.
Because analytic gradients are unavailable, we approximate 
\(\nabla_{\mathbf{u}}E_{\text{phys}}\) by forward finite differences.
Let \(E(\mathbf{u})=E_{\text{phys}}(P,M(\mathbf{u}))\).
For step size \(\epsilon\),
\begin{equation}
\frac{\partial E}{\partial u_i} \approx 
\frac{E(\mathbf{u}+\epsilon\mathbf{e}_i)-E(\mathbf{u})}{\epsilon}.
\label{eq:finite_diff}
\end{equation}
We set \(\epsilon=10^{-3}\) in all runs after scale normalisation of \(\mathbf{u}\).

{
\setlength{\intextsep}{0pt}        
\setlength{\abovecaptionskip}{-8pt} 
\begin{figure*}[ht]
  \centering
  \begin{subfigure}[b]{0.32\textwidth}
    \centering
    \includegraphics[width=\linewidth]{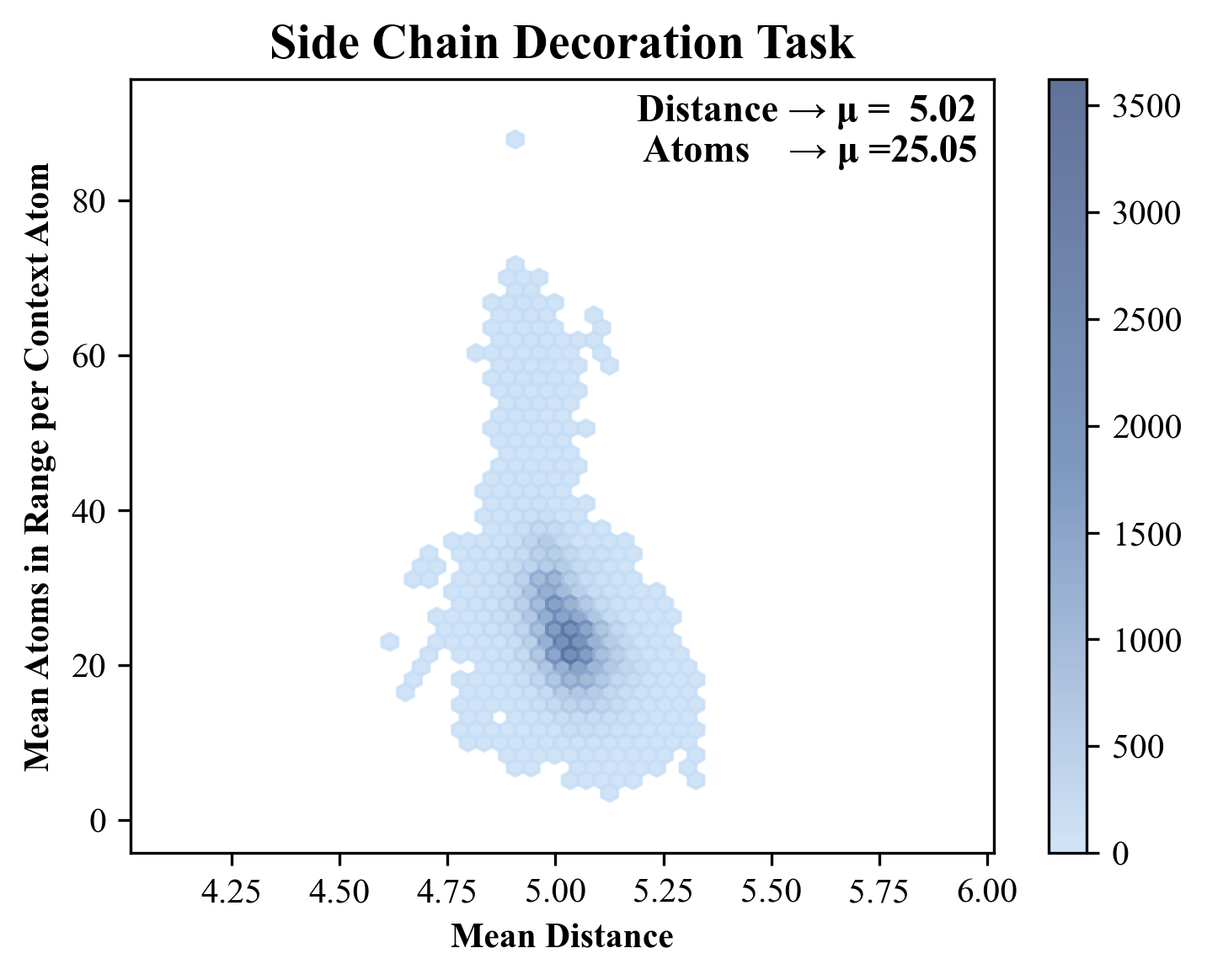}
    \label{fig:sub:a}
  \end{subfigure}
  \hfill
  \begin{subfigure}[b]{0.32\textwidth}
    \centering
    \includegraphics[width=\linewidth]{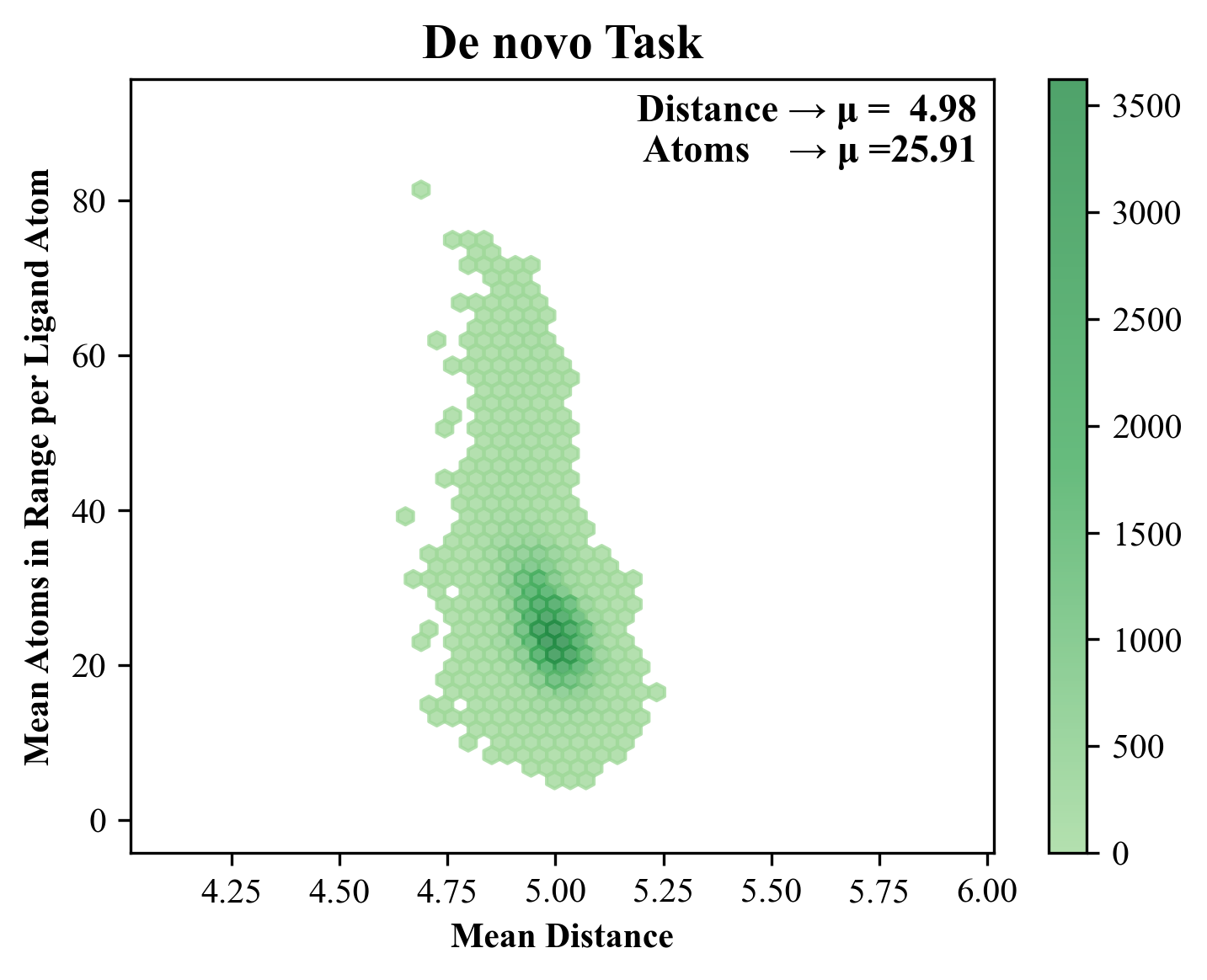}
    \label{fig:sub:b}
  \end{subfigure}
  \hfill
  \begin{subfigure}[b]{0.3\textwidth}
    \centering
    \includegraphics[width=\linewidth]{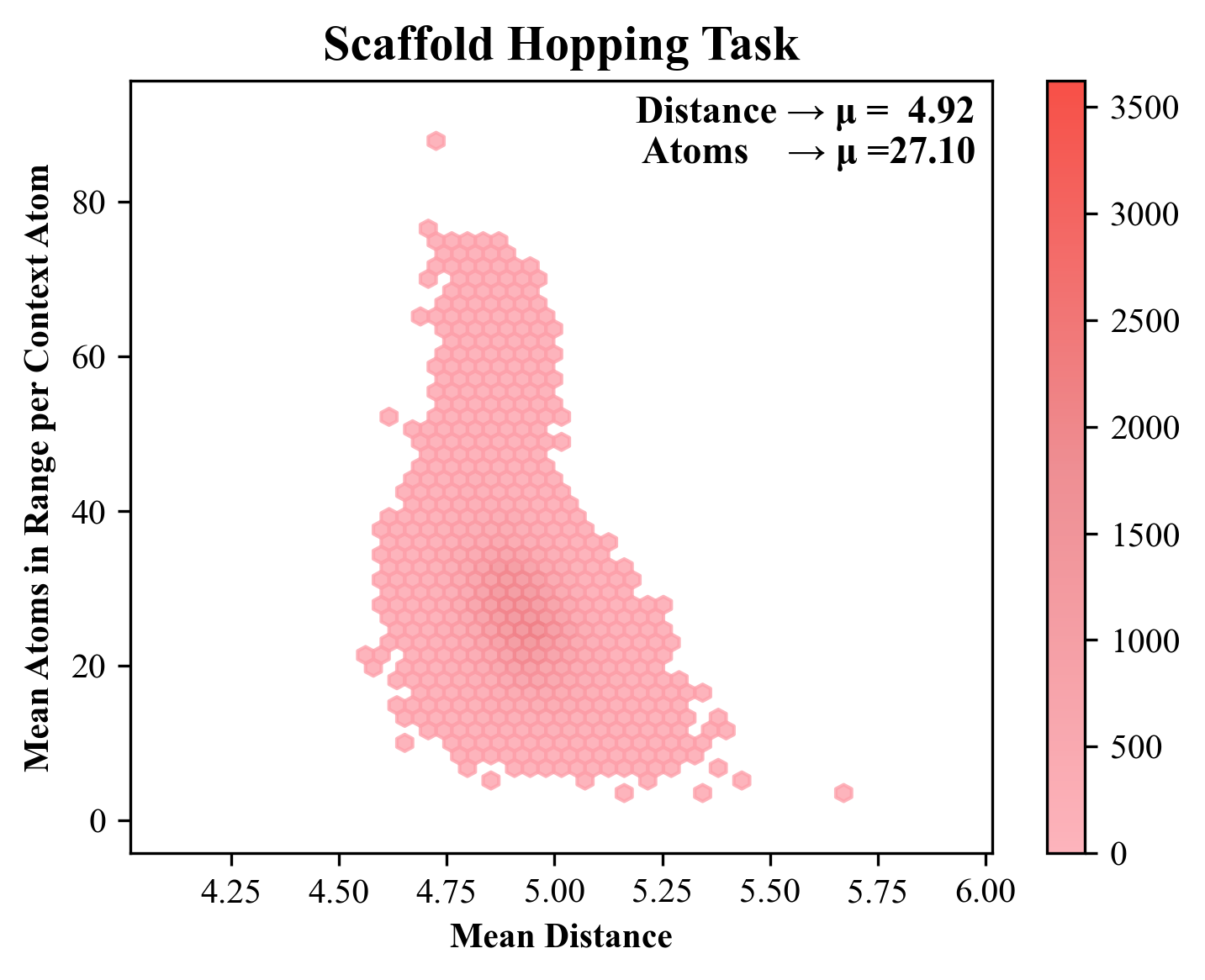}
    \label{fig:sub:c}
  \end{subfigure}
  \hfill
  \setlength{\lineskiplimit}{0pt}
  \setlength{\lineskip}{-12pt}  
  \begin{subfigure}[b]{0.32\textwidth}
    \centering
    \includegraphics[width=\linewidth]{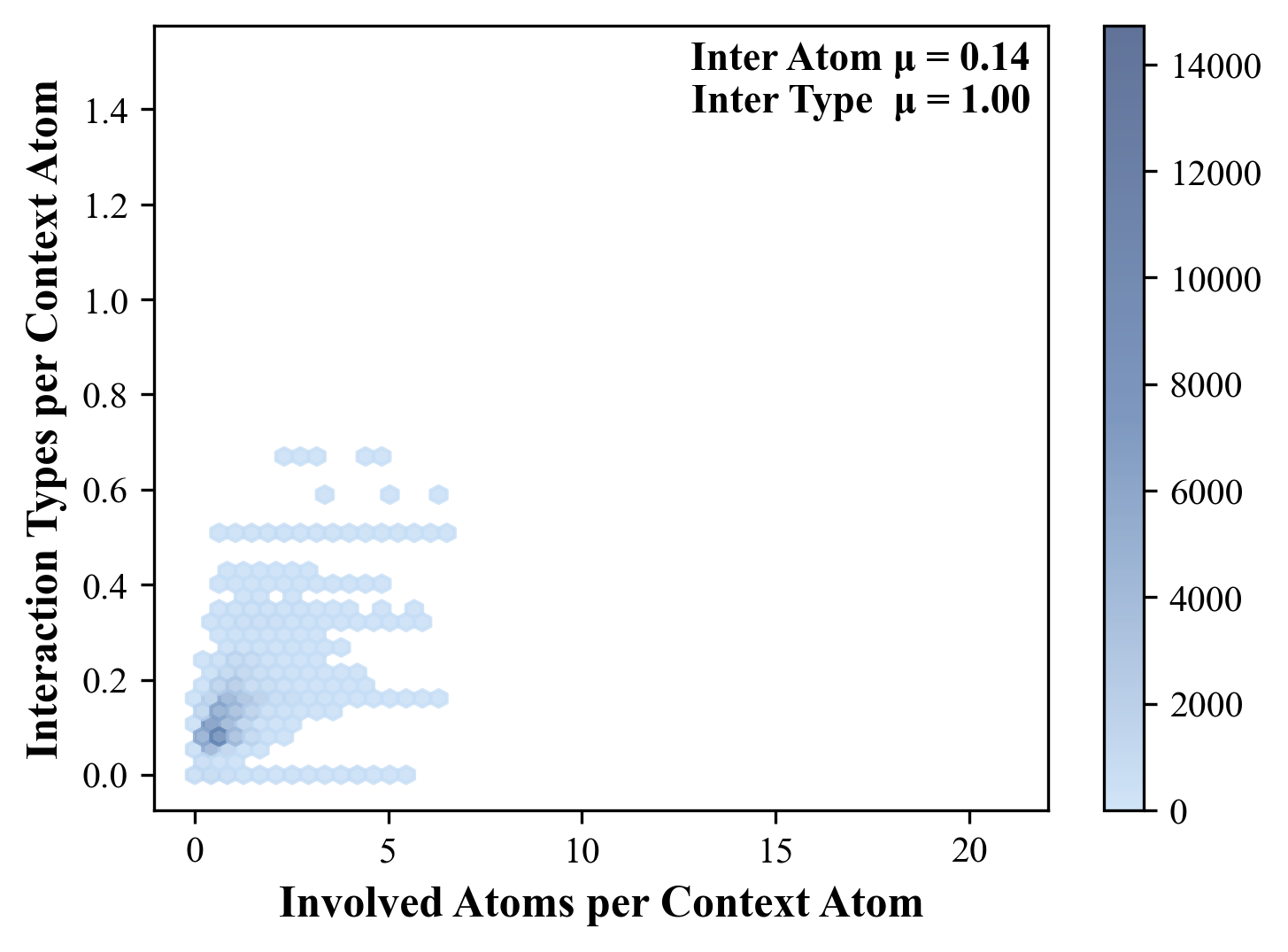}
    \label{fig:sub:d}
  \end{subfigure}
  \hfill
  \begin{subfigure}[b]{0.3\textwidth}
    \centering
    \includegraphics[width=\linewidth]{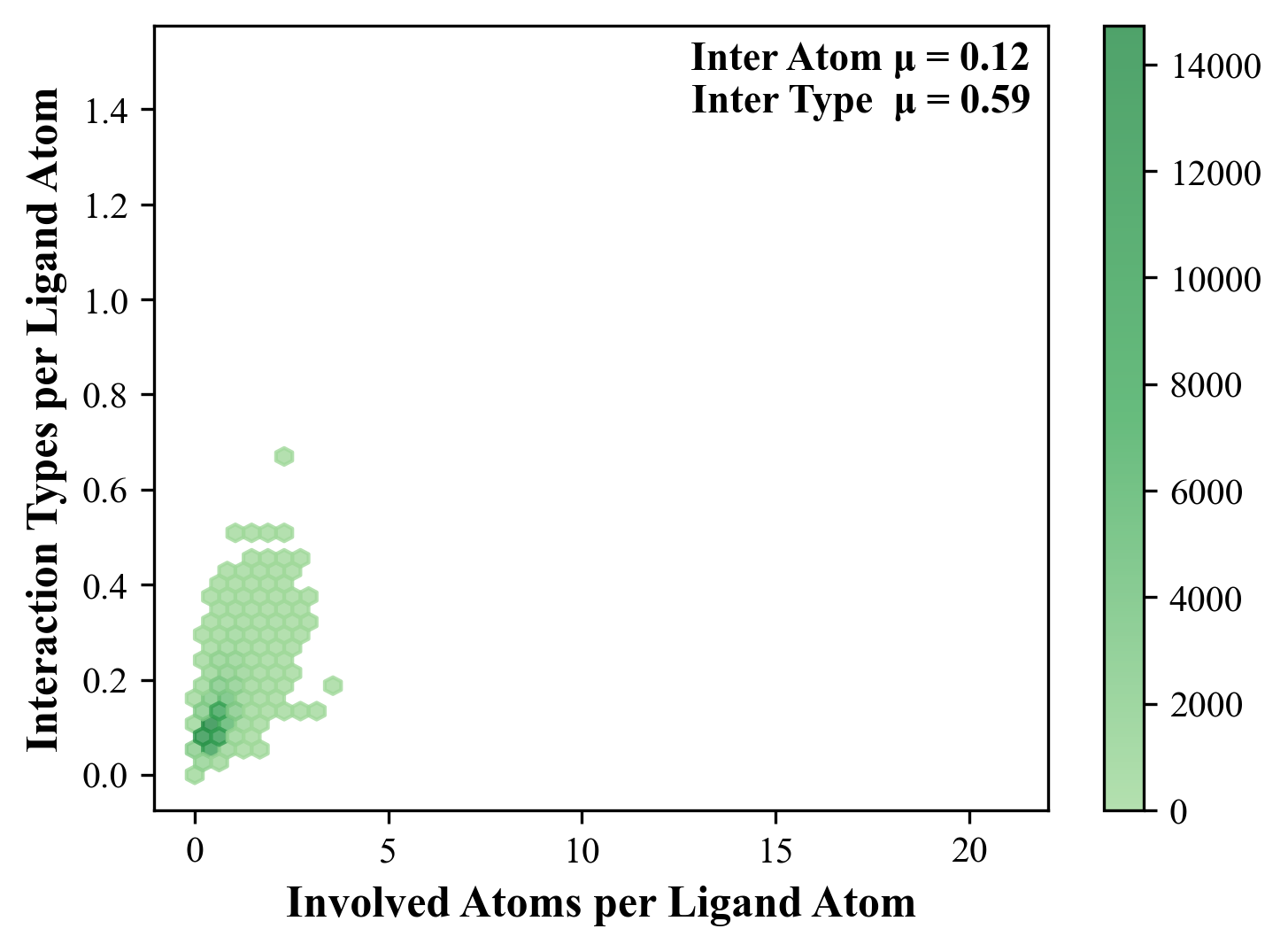}
    \label{fig:sub:e}
  \end{subfigure}
  \hfill
  \begin{subfigure}[b]{0.3\textwidth}
    \centering
    \includegraphics[width=\linewidth]{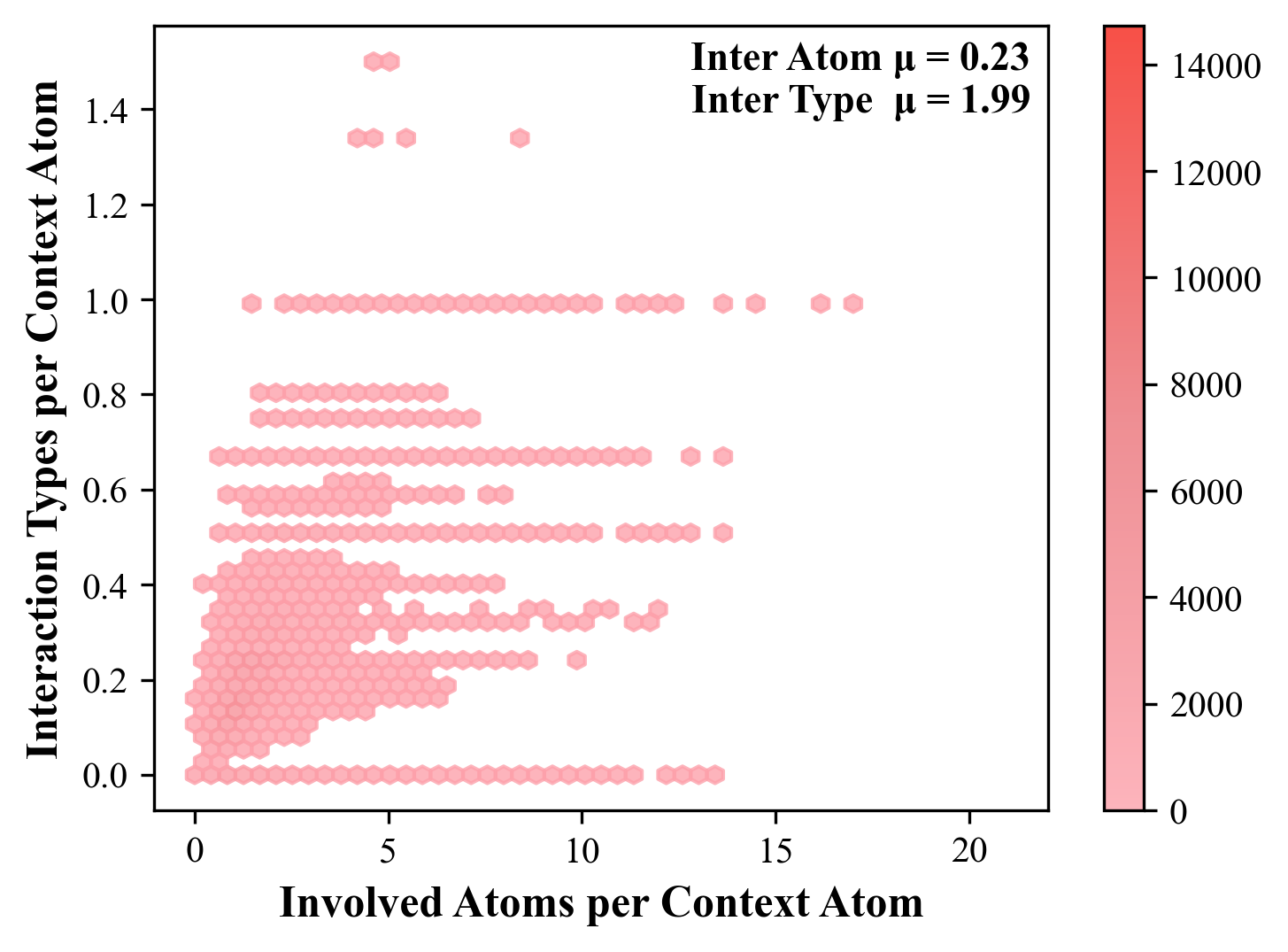}
    \label{fig:sub:f}
  \end{subfigure}
\caption{Hexagon–bin density maps for SC, DN, and SH. 
Top panels: mean edge length (\,\(\bar d\)\,) vs.\ mean number of neighbours per context atom (\,\(\bar n\)\,). 
Bottom panels: mean number of interacting atoms (\,\(\bar k\)\,) vs.\ mean number of interaction types (\,\(\bar t\)\,). 
DN values are averaged over all ligand atoms. 
Insets show task-level means. SH spans the broadest range and attains the highest means on all four axes, indicating richer geometric and chemical context than DN and SC.}
  \label{fig:hexbin}
\end{figure*}
}

We run a Limited-memory BFGS search optimiser with fixed learning rate (0.1)~\cite{KarmaDock}.
At iteration \(k\):
\begin{enumerate}
  \item Evaluate \(E(\mathbf{u}_k)\) and its finite-difference gradient.
  \item Perform an L-BFGS update to propose \(\mathbf{u}_{k+1}\).
  \item Update the ligand pose; recompute the energy.
  \item Track the best energy so far.
\end{enumerate}
We run at most \(T_{\max}\) iterations (\(T_{\max}= \) epochs command-line argument).
The initial \(\mathbf{u}_0=\mathbf{0}\) uses the generator pose.
Let \(E_{\text{init}}\) and \(E_{\text{opt}}\) be the energies before and after refinement.
If \(E_{\text{opt}} \le E_{\text{init}}\) we accept the refined pose; otherwise we keep the initial one.
Both the kept structure and the tracked scores are saved for later analysis.
This rule prevents noisy gradients from degrading good initial placements.

\section{Results}
\label{Sec:results}

\begin{figure}[ht]
    \setlength{\abovecaptionskip}{-4pt} 
    \centering
    \includegraphics[width=0.45\textwidth]{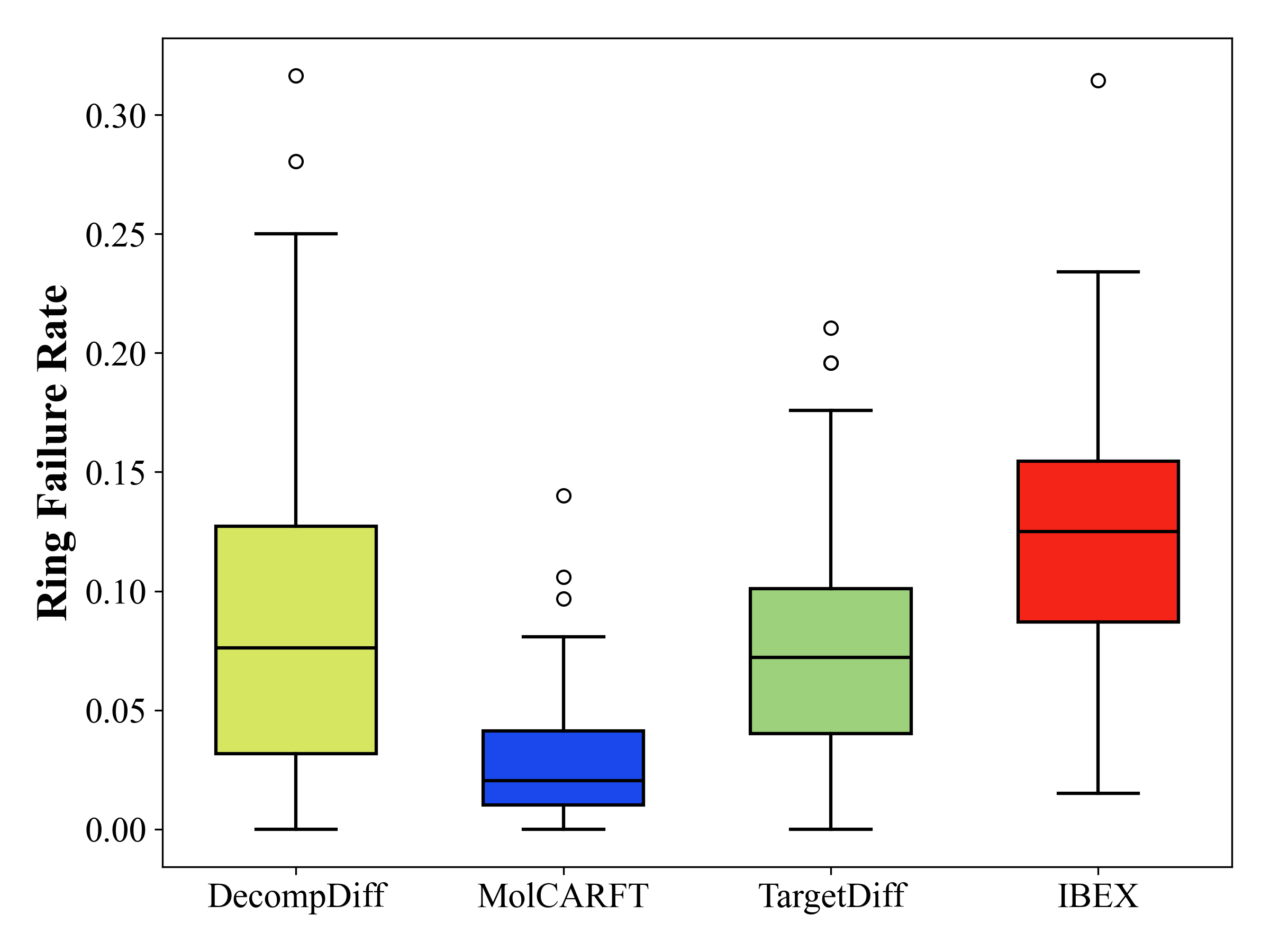} 
    \caption{Harder geometry implies higher \(I_m\), supporting the ordering above. IBEX exhibits the highest failure rate, followed by DecompDiff, TargetDiff, and MolCRAFT.} 
    \label{Fig:geom}
\end{figure}

CrossDocked2020~\cite{crossdocked2020} is one of the most widely used benchmarks for structure-based drug design, providing paired three-dimensional structures of protein pockets and docked ligands. 
Existing methods have adopted different data splits and evaluation protocols. 
CBGBench~\cite{lin2024cbgbench} follows the split defined by LiGAN~\cite{LiGAN} and 3DSBDD~\cite{3DSBDD} and prevents label leakage by constructing the side chain and scaffold tasks only after an independent train/test partition. 
Models are evaluated from four complementary perspectives: interaction quality, chemical properties, geometric accuracy, and substructure validity. 
The benchmark integrates a diverse panel of state-of-the-art generators, including LiGAN, 3DSBDD, VoxBind~\cite{VoxBind}, diffusion models (TargetDiff~\cite{targetdiff}, DiffSBDD~\cite{diffsbdd}, DecompDiff~\cite{decompdiff}, DiffBP~\cite{diffbp}, D3FG~\cite{d3fg}, MolCRAFT~\cite{molcraft}), and autoregressive models (Pocket2Mol~\cite{pocket2mol}, GraphBP~\cite{graphbp}, FLAG~\cite{flag}). (Refer to Appendix E.)

{
\setlength{\abovecaptionskip}{0pt} 
\setlength{\belowcaptionskip}{2pt} 
\begin{figure*}[!t]
  \centering
  \begin{subfigure}[!t]{0.24\textwidth}
  \setlength{\abovecaptionskip}{-2pt} 
    \centering
    \includegraphics[width=\linewidth]{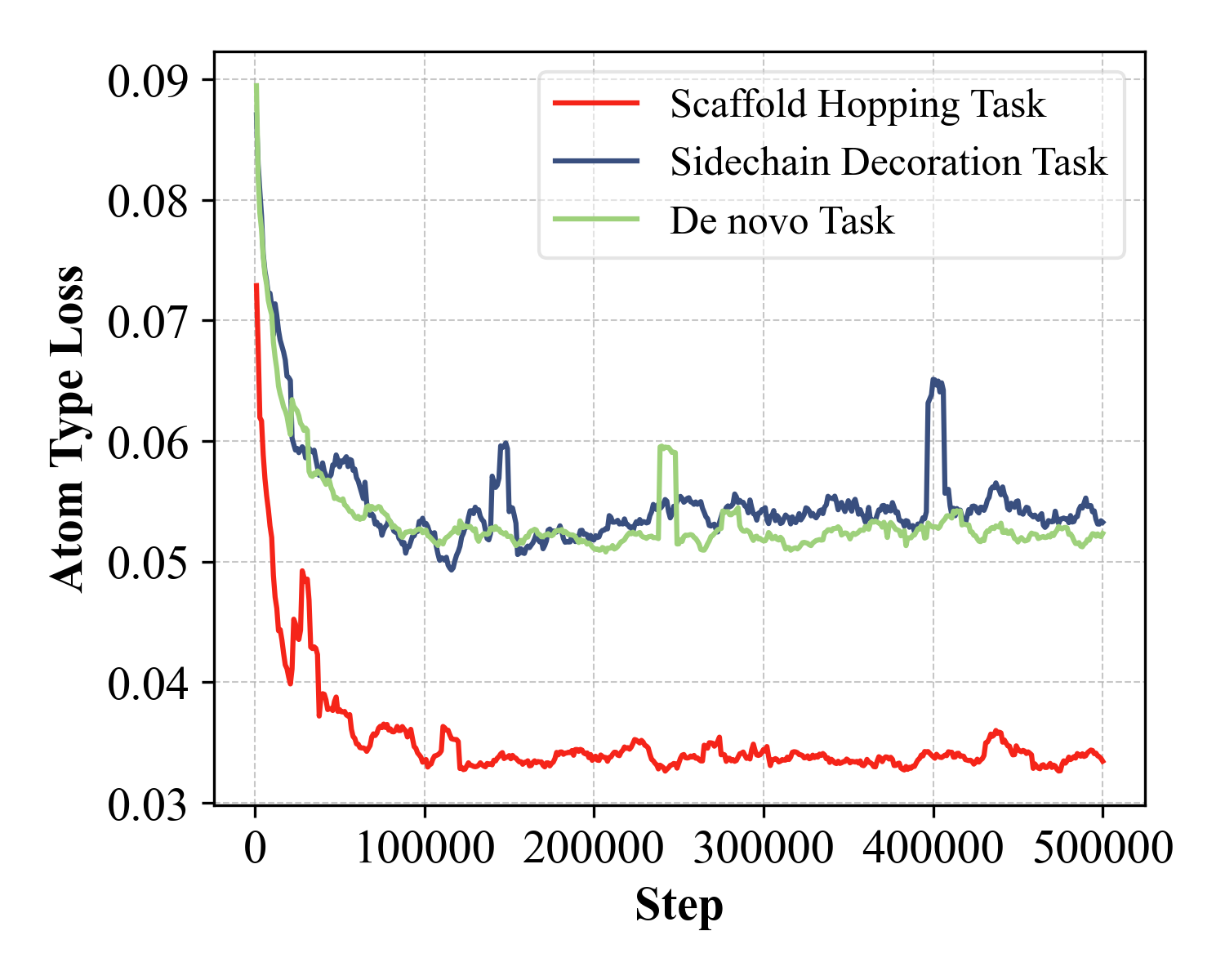}
    \caption{Atom‑type classification loss}
    \label{fig:sub_atom_loss}
  \end{subfigure}
  \hfill
  \begin{subfigure}[!t]{0.24\textwidth}
  \setlength{\abovecaptionskip}{-2pt} 
    \centering
    \includegraphics[width=\linewidth]{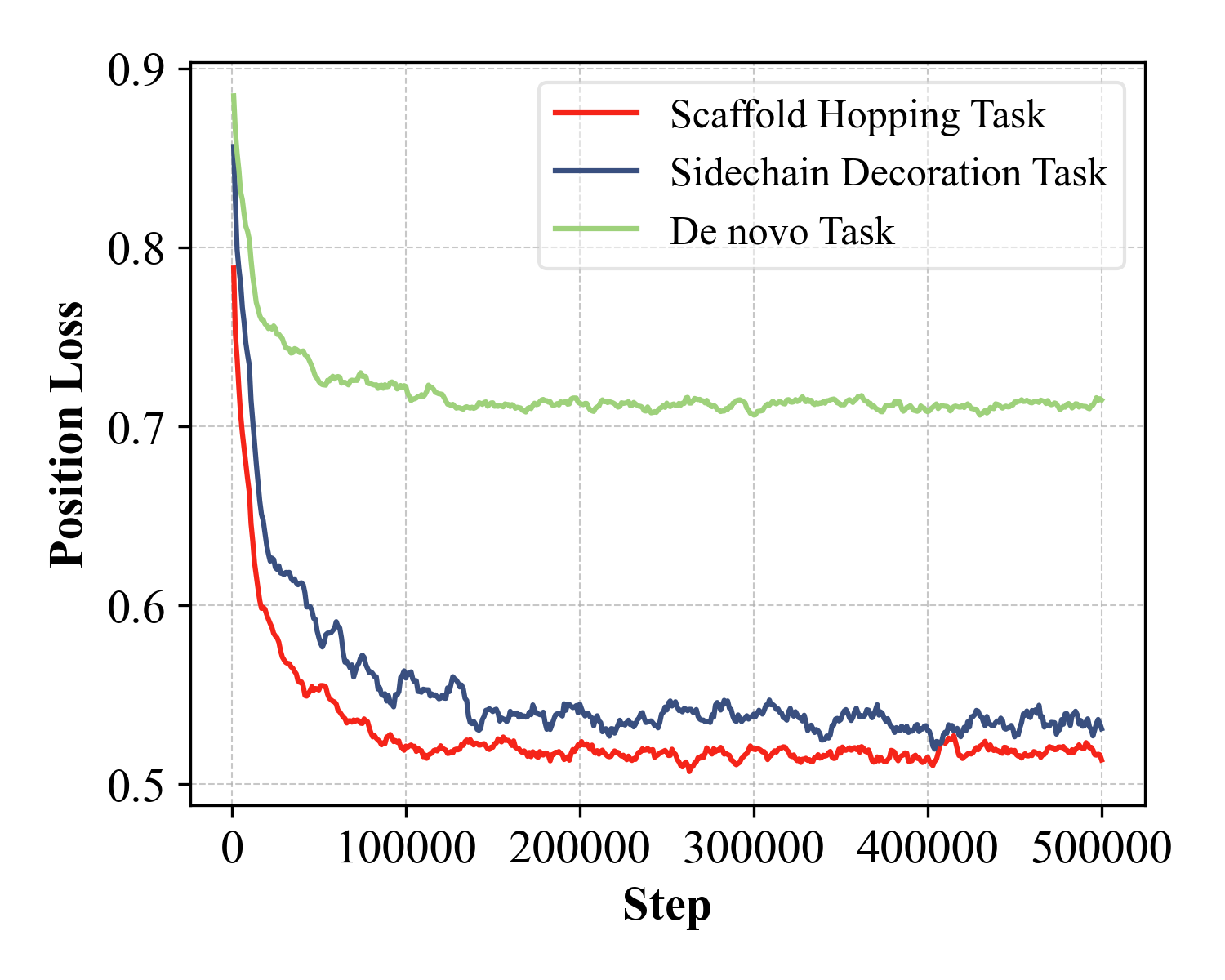}
    \caption{Position‑reconstruction loss}
    \label{fig:sub_pos_loss}
  \end{subfigure}
  \hfill
  \begin{subfigure}[!t]{0.24\textwidth}
  \setlength{\abovecaptionskip}{-2pt} 
    \centering
    \includegraphics[width=\linewidth]{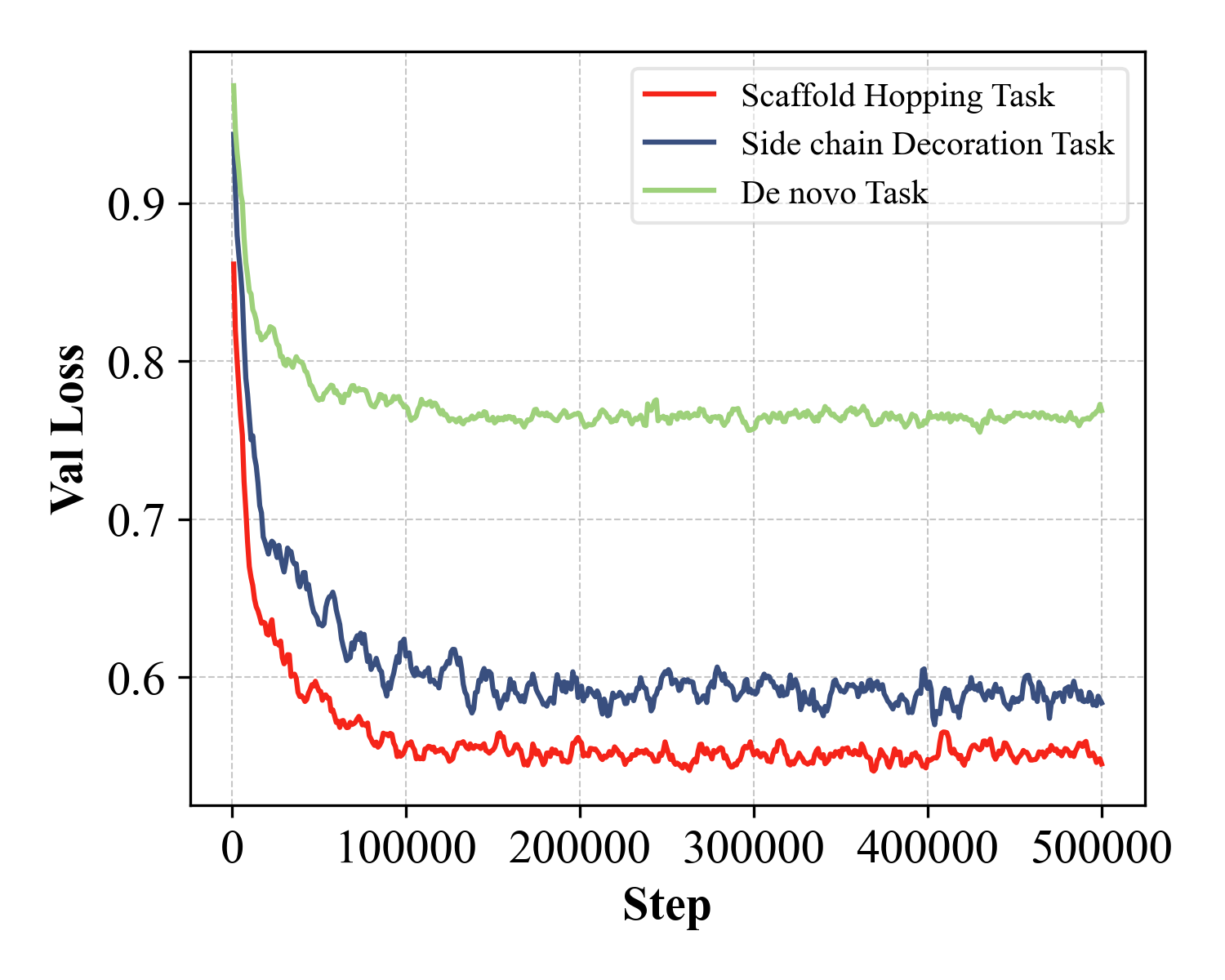}
    \caption{Validation total loss}
    \label{fig:sub_val_loss}
  \end{subfigure}
  \hfill
  \begin{subfigure}[!t]{0.24\textwidth}
  \setlength{\abovecaptionskip}{-2pt} 
    \centering
    \includegraphics[width=\linewidth]{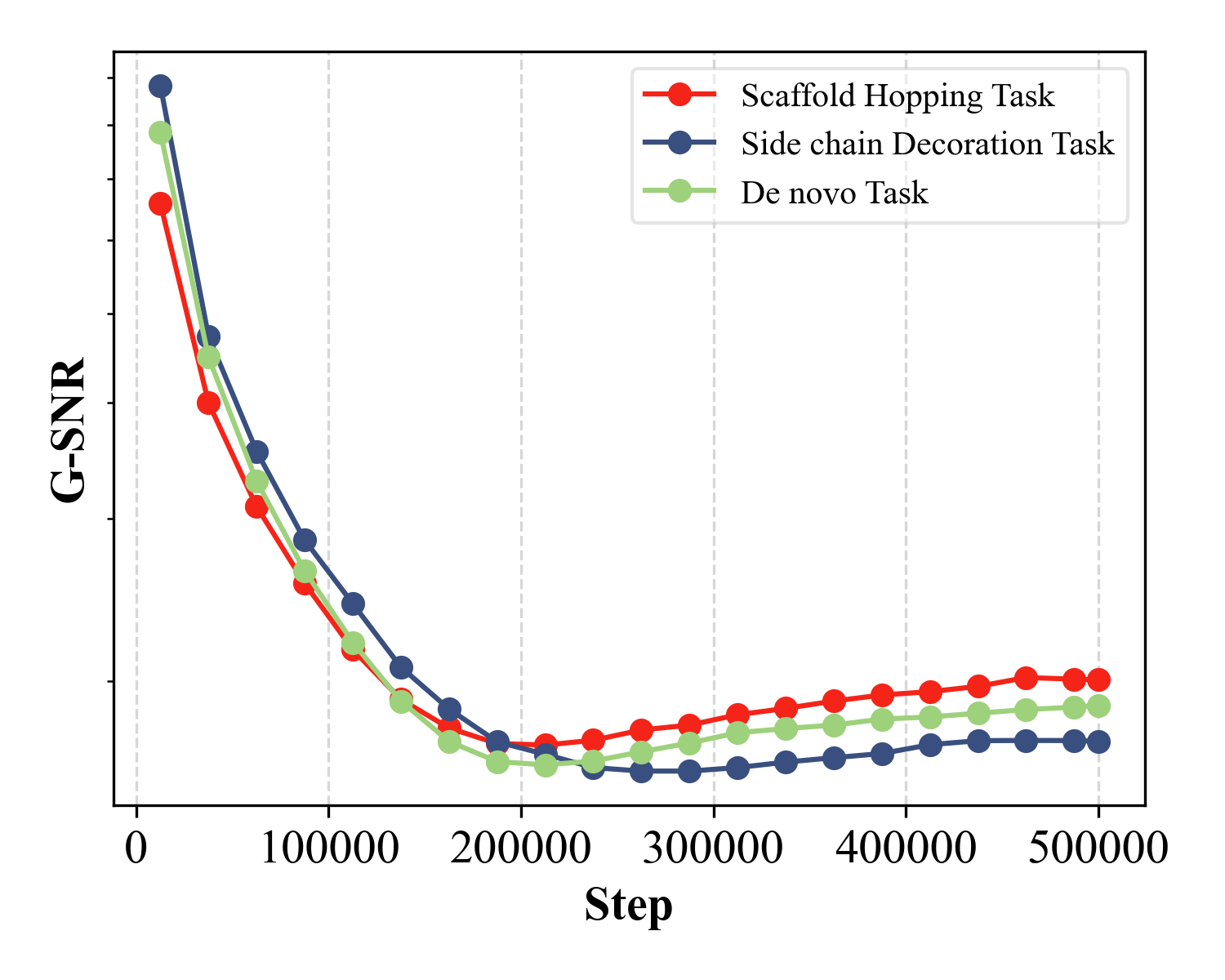}
    \caption{Gradient-SNR}
    \label{fig:sub_gsnr}
  \end{subfigure}
  \caption{Training and validation dynamics of IBEX models on three generation tasks. Panels show atom‑type classification loss, position‑reconstruction loss, total validation loss, and gradient signal‑to‑noise ratio (G‑SNR) as functions of training steps for the Scaffold Hopping (red), Side‑chain Decoration (blue), and De novo (green) tasks}
  \label{fig:loss}
\end{figure*}
}

\begin{figure*}[ht]
\setlength{\abovecaptionskip}{-8pt} 
  \centering
  \begin{subfigure}[b]{0.32\textwidth}
    \centering
    \includegraphics[width=\linewidth]{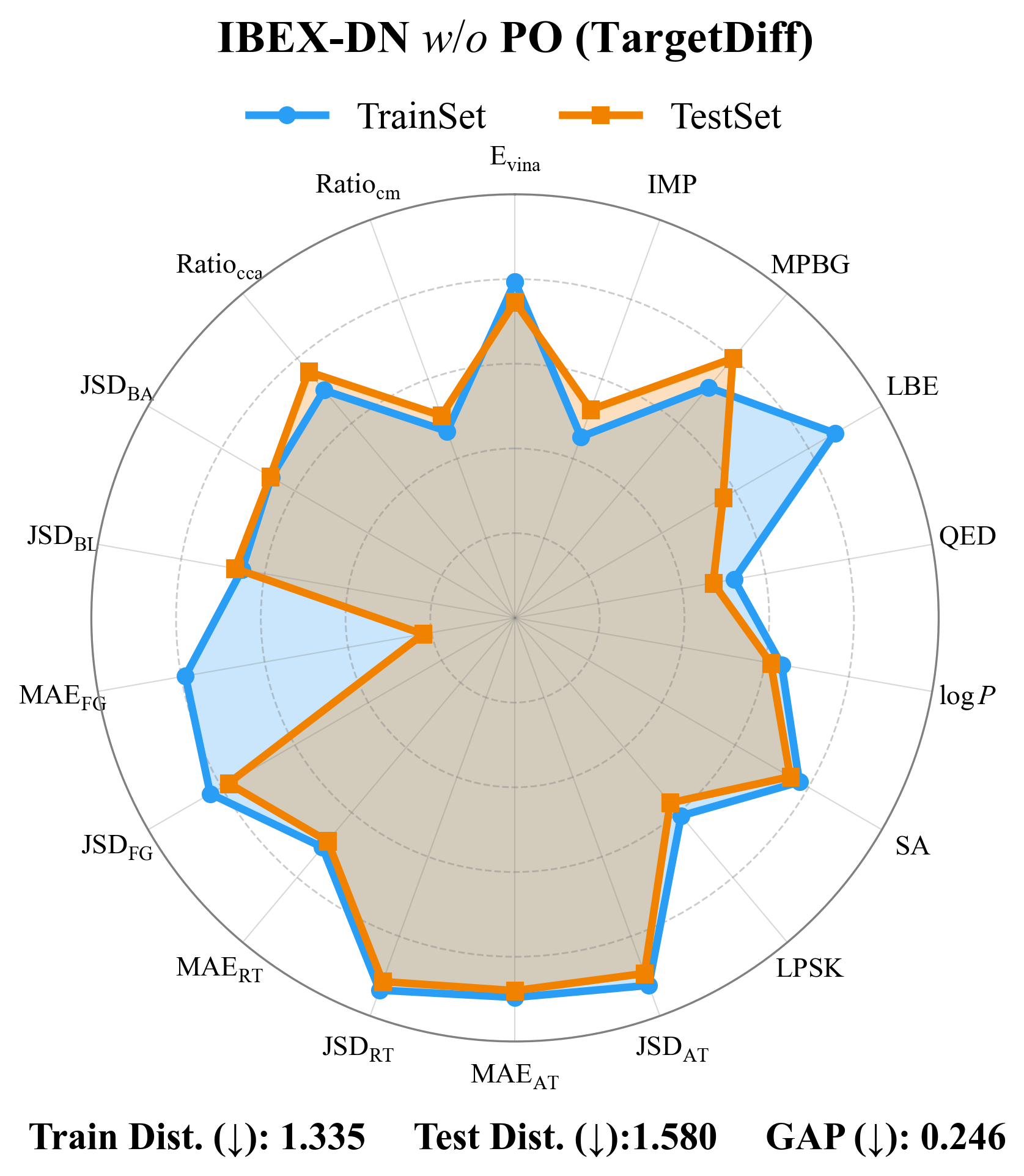}
    \label{fig:sub_radar_dn}
  \end{subfigure}
  \hfill
  \begin{subfigure}[b]{0.32\textwidth}
    \centering
    \includegraphics[width=\linewidth]{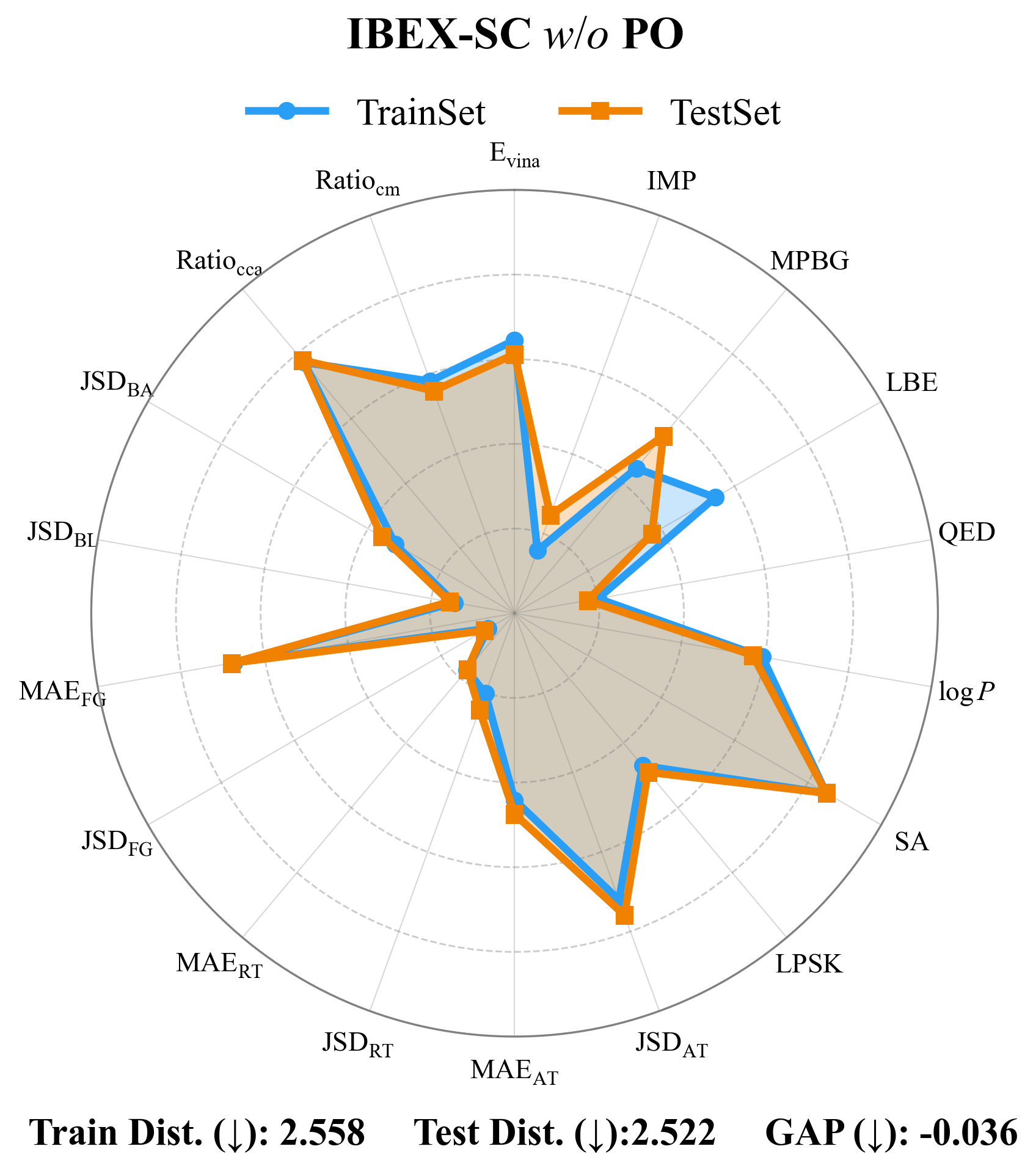}
    \label{fig:sub_radar_sc}
  \end{subfigure}
  \hfill
  \begin{subfigure}[b]{0.32\textwidth}
    \centering
    \includegraphics[width=\linewidth]{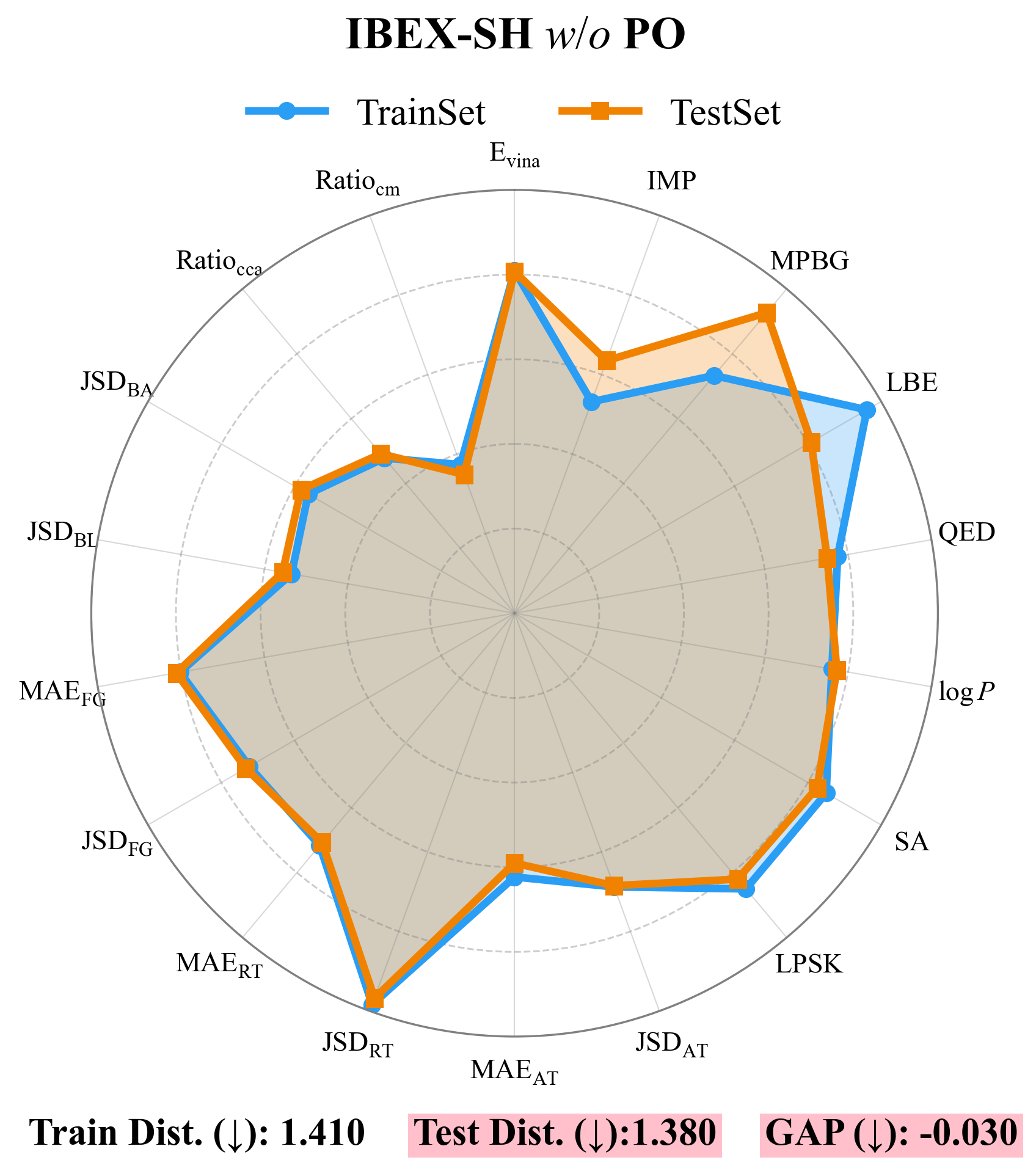}
    \label{fig:sub_radar_sh}
  \end{subfigure}
\caption{Train–test comparison on 18 normalised metrics for three IBEX settings. 
For each IBEX variant we track \emph{eighteen} chemistry-aware metrics covering binding energy, physicochemical properties, geometric complementarity, and distribution alignment. 
Every metric \(k\) is min–max normalised to \([0,1]\) (monotonic in the beneficial direction), and the profile distance is the Euclidean norm \(\lVert\mathbf{1}-\hat{\mathbf v}\rVert_2\) between a sample’s normalised vector \(\hat{\mathbf v}\in[0,1]^{18}\) and the ideal all-ones target. 
The radar overlays Train (blue) and Test (orange) envelopes; the number beneath each chart reports the test distance, and GAP is the absolute train–test difference. 
The SH task exhibits the tightest Train–Test overlap, the smallest profile distance on the held-out set, and the narrowest GAP, illustrating how increased task difficulty promotes robust generalization.
}
  \label{fig:radar}
\end{figure*}

\textbf{Task Difficulty under Geometric Constraints.}
Matched ligand sets were generated with four diffusion baseline: 
IBEX (SH), DecompDiff (Scaffold-Arms), 
TargetDiff and MolCRAFT (no geometric constraint). 
Each output was screened by RDKit~\cite{rdkit} topology checks to detect unclosed rings. A higher failure rate signals a harder but more informative task~\cite{jiang2024pocketflow}. 
SH task exposes the network to explicit side chain–pocket interactions during training but leaves these atoms un-denoised; at test time, the model must denoise them from scratch, increasing conflict yet enriching the learned representation. 
This supports our claim that SH operates in the high-information regime. \textbf{Location beats quantity.} Retaining side chain context boosts mutual information and thus effective capacity, outweighing a lower overall mask ratio.

{
\begin{table*}[ht]
  \centering
  \fontsize{8}{10}\selectfont
  \setlength{\tabcolsep}{2pt}
  \renewcommand{\arraystretch}{1.0}
\begin{tabular}{l|cc|cr|cc|rcrr|cccc|ccccc}
\toprule
\multicolumn{1}{c|}{Model} 
  & \multicolumn{2}{c|}{Ablation} 
  & \multicolumn{2}{c|}{Vina Score} 
  & \multicolumn{2}{c|}{Vina Min} 
  & \multicolumn{4}{c|}{Vina Dock} 
  & \multicolumn{4}{c|}{PLIP Interaction}
  & \multicolumn{5}{c} {Chemical property}\\
\cmidrule(lr){2-3} \cmidrule(lr){4-5} \cmidrule(lr){6-7} \cmidrule(lr){8-11} \cmidrule(lr){12-15} \cmidrule(lr){16-20}
  & SH & PR
  & $\mathrm{E}_{\mathrm{vina}}$ & IMP
  & $\mathrm{E}_{\mathrm{vina}}$ & IMP 
  & $\mathrm{E}_{\mathrm{vina}}$ & IMP & MPBG & LBE  
  & $\mathrm{MAE}_{\mathrm{OA}}$    & $\mathrm{JSD}_{\mathrm{OA}}$  
  & $\mathrm{MAE}_{\mathrm{PP}}$    & $\mathrm{JSD}_{\mathrm{PP}}$ 
  & QED & LogP & SA & LPSK & Validity\\
\midrule
LiGAN & - & -  & \textbf{-6.47} & \textbf{62.13} & \textbf{-7.14} & \textbf{70.18} & \underline{-7.70} & \textbf{72.71} &   4.22 & \underline{0.3897} & 0.0905 & 0.0346 & \textbf{0.3416} & \underline{0.1451} & 0.46 &  \textbf{0.56} & \textbf{0.66} & 4.39 & 0.42\\
3DSBDD & - & - & -     &  3.99 & -3.75 & 17.98 & -6.45 & 31.46 &   9.18 & 0.3839 & 0.0934 & 0.0392 & 0.4231 & 0.1733 & 0.48 &  \textbf{0.47} & 0.63 & 4.72 & 0.54\\
GraphBP       & - & - & -     &  0.00 &  -    &  1.67 & -4.57 & 10.86 & -30.03 & 0.3200 & 0.1625 & 0.0462 & 0.4835 & 0.2101 & 0.44 &  \textbf{3.29} & 0.64 & 4.73 & 0.66\\
Pocket2mol    & - & - & -5.23 & 31.06 & -6.03 & 38.04 & -7.05 & 48.07 &  -0.17 & \textbf{0.4115} & 0.2455 & 0.0319 & \underline{0.4152} & \underline{0.1535} & 0.39 &  \textbf{2.39} & 0.65 & 4.58 & 0.75 \\
TargetDiff    & - & - & -5.71 & 38.21 & -6.43 & 47.09 & -7.41 & 51.99 &   5.38 & 0.3537 & 0.0600 & \underline{0.0198} & 0.4687 & 0.1757 & \underline{0.49} &  \textbf{1.13} & 0.60 & 4.57 & \textbf{0.96} \\
DiffSBDD      & - & - & -     & 12.67 & -2.15 & 22.24 & -5.53 & 29.76 & -23.51 & 0.2920 & 0.1461 & 0.0333 & 0.5265 & 0.1777 & \underline{0.49} & \textbf{-0.15} & 0.34 & \underline{4.89}  & 0.71 \\
DiffBP        & - & - & -     &  8.60 &  -    & 19.68 & -7.34 & 49.24 &   6.23 & 0.3481 & 0.1430 & 0.0249 & 0.5639 & \textbf{0.1256} & 0.47 &  \textbf{5.27} & 0.59 & 4.47 & 0.78\\
FLAG          & - & - & -     &  0.04 & -     &  3.44 & -3.65 & 11.78 & -47.64 & 0.3319 & \underline{0.0277} & \textbf{0.0170} & \underline{0.3976} & 0.2762 & 0.41 &  \textbf{0.29} & 0.58 & \textbf{4.93} & 0.68 \\
D3FG          & - & - & -     &  3.70 & -2.59 & 11.13 & -6.78 & 28.90 &  -8.85 & \underline{0.4009} & \textbf{0.0135} & 0.0638 & 0.4641 & 0.1850 & \underline{0.49} &  \textbf{1.56} & \underline{0.66} & \underline{4.84} & 0.77\\ 
DecompDiff    & - & - & -5.18 & 19.66 & -6.04 & 34.84 & -7.10 & 48.31 &  -1.59 & 0.3460 & 0.0769 & 0.0215 & 0.4369 & 0.1848 & \underline{0.49} &  \textbf{1.22} & \underline{0.66} & 4.40  & 0.89 \\
MolCARFT      & - & - & \underline{-6.15} & \underline{54.25} & \underline{-6.99} & \underline{56.43} & \underline{-7.79} & \underline{56.22} &   \underline{8.38} & 0.3638 & 0.0780 & 0.0214 & 0.4574 & 0.1868 & 0.48 &  \textbf{0.87} & \underline{0.66} & 4.39 & \underline{0.95} \\
VoxBind       & - & - & \underline{-6.16} & \underline{41.80} & \underline{-6.82} & \underline{50.02} & -7.68 & 52.91 &   \underline{9.89} & 0.3588 & \underline{0.0533} & 0.0257 & 0.4606 & 0.1850 & \underline{0.54} &  \textbf{2.22} & 0.65 & 4.70 & 0.74\\
IBEX  & \cmark & \cmark & -3.09 & 37.67 & -5.23 & 47.34 & \textbf{-8.09} & \underline{63.69} &  \textbf{14.69} & 0.3813 & 0.0709 & \underline{0.0176} & 0.4670 & 0.1947 & \textbf{0.60} &  \textbf{2.73} & 0.63 & 4.82 & \textbf{0.96} \\ 
\midrule
IBEX-DN  & \xmark & \xmark & \textbf{-5.71} & \textbf{38.21} & \textbf{-6.43} & 47.09 & -7.41 & 51.99 &   5.38 & 0.3537 & 0.0600 & 0.0198 & 0.4687 & 0.1757 & 0.49 &  \textbf{1.13} & 0.60 & 4.57 & \textbf{0.96} \\
IBEX-SC & \xmark & \xmark  & -3.53 & 18.54 & -4.73 & 21.89 & -6.20 & 24.81 & -10.22 & 0.3416 & \textbf{0.0430} & 0.5696 & 0.4801 & \textbf{0.0263} & 0.35 & \textbf{0.85} & \textbf{0.63} & 4.38 & 0.54 \\
IBEX-SH  & \cmark & \xmark & -1.96 & 31.03 & -5.06 & 46.58 & -8.07 & 63.50 &  \textbf{14.87} & 0.3809 & 0.0698 & 0.0198 & 0.5442 & 0.1897 & \textbf{0.60} &  \textbf{2.73} & \textbf{0.63} & \textbf{4.82} & \textbf{0.96} \\
IBEX  & \cmark & \cmark & -3.09 & 37.67 & -5.23 & \textbf{47.34} & \textbf{-8.09} & \textbf{63.69} &  14.69 & \textbf{0.3813} & 0.0709 & \textbf{0.0176} & \textbf{0.4670} & 0.1947 & \textbf{0.60} &  \textbf{2.73} & \textbf{0.63} & \textbf{4.82} & \textbf{0.96} \\ 
\bottomrule
\end{tabular}
\caption{Aggregate docking, interaction, and physicochemical metrics for recent generative pipelines}
  \label{tab:combined_metrics1}
\end{table*}
}

{
\setlength{\abovecaptionskip}{-2pt} 
\begin{figure*}[ht]       
\centering
\includegraphics[width=\linewidth]{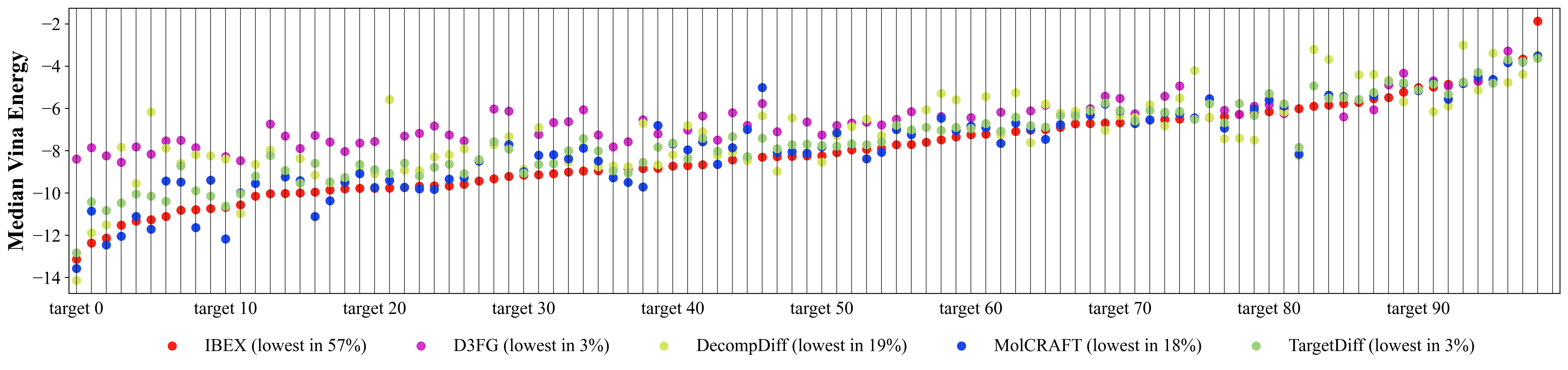}
\caption{Per‑target docking performance of five generative pipelines on 100 held‑out CBGBench (same as CrossDocked2020) receptors. Each coloured point is the median AutoDock Vina binding energy of the candidate set produced for one target by IBEX, D3FG, DecompDiff, MolCRAFT, and TargetDiff.}
\label{fig:dotmap}
\end{figure*}
}

{
\setlength{\abovecaptionskip}{0pt} 
\begin{figure*}[!t]       
\centering
\includegraphics[width=0.98\textwidth]{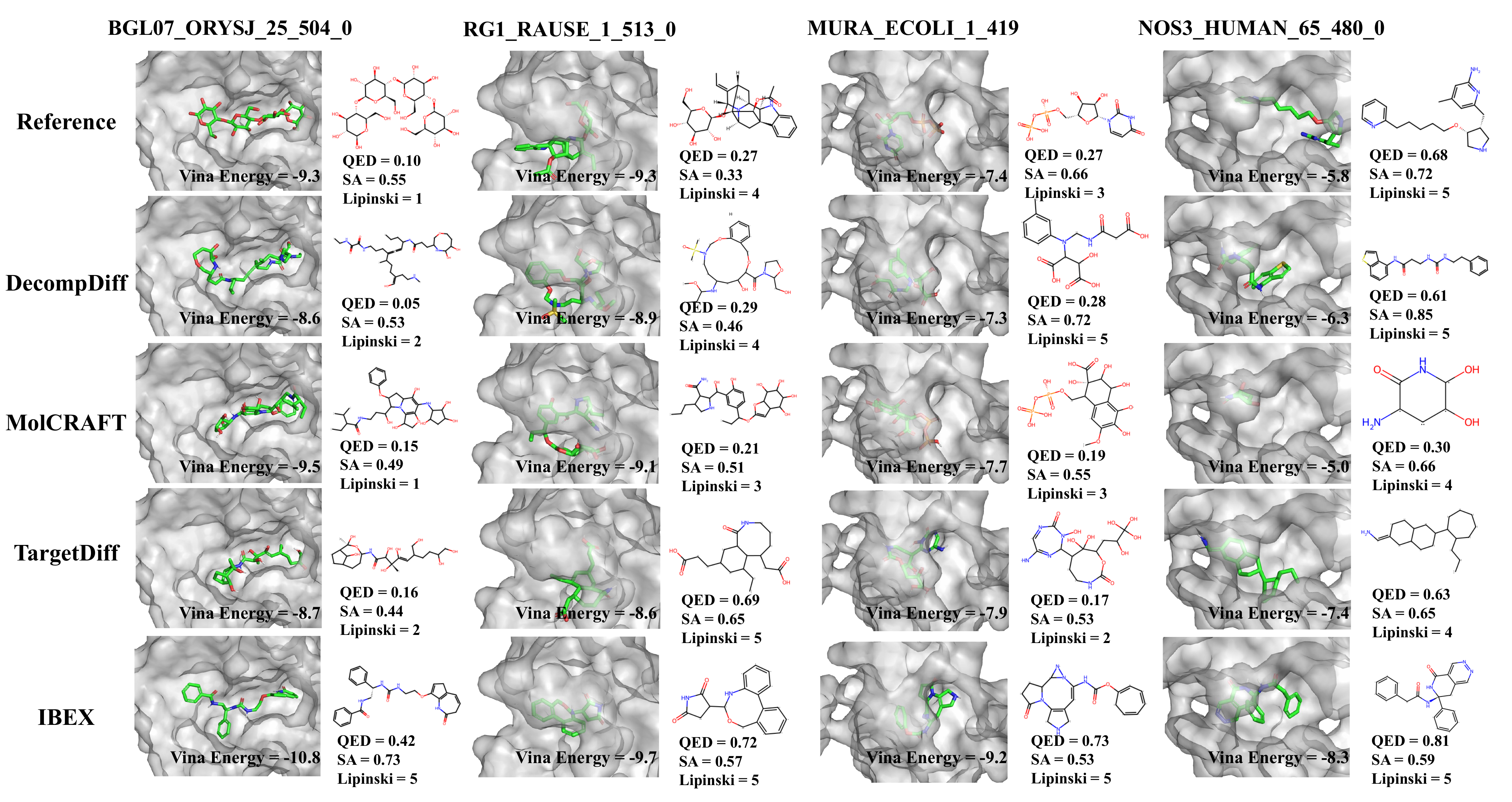}
\caption{Docking poses and drug-likeness of median-score candidates on four CBGBench test pockets. Rows list the crystal reference and the \emph{median‐ranked} molecule (by AutoDock Vina) from DecompDiff, MolCRAFT, TargetDiff, and IBEX. Each ligand is depicted with its predicted pose (green sticks) and associated Vina energy, alongside its planar formula annotated with QED, SA, and Lipinski compliance.}
\label{fig:case}
\end{figure*}
}

\textbf{Task-dependent Information-Bottleneck Analysis.}
For each protein–ligand complex, we connect every pair of heavy atoms whose Euclidean separation is below
\(6\,\text{\AA}\) and treat the resulting links as virtual edges~\cite{diffdock, unimol}.
The complex is then compressed into a four-dimensional summary
\(Z=(\bar n,\,\bar d,\,\bar t,\,\bar k)\) that lies on two orthogonal planes: \textbf{Distance plane.}
(a) Neighbour density \(\bar n\): the mean number of atoms found within the sphere of each context atom;  
(b) Mean edge length \(\bar d\): the average virtual-edge distance. These two axes quantify, respectively, the strength of short-range physical electrostatic forces. \textbf{Interaction plane.} (c) Type richness \(\bar t\): the mean count of distinct interaction categories—hydrophobic, hydrogen bond, water-bridge, \(\pi\)–\(\pi\) stack, \(\pi\)–cation, halogen, metal—triggered by a context atom~\cite{plip}; (d) Contact multiplicity \(\bar k\): the mean number of protein atoms that realise those interactions. These two metrics capture the chemical complexity and the interaction strength~\cite{ipdiff}. For DN generation, the model possesses no ligand-derived context atoms; all four statistics are therefore computed over every ligand atoms. SH and SC restrict the tally to a predefined context subset, leading to visibly broader hex-bin distributions for SH in both planes (Fig.~\ref{fig:hexbin}). Under the PAC-Bayes information-bottleneck framework\cite{PAC-Bayes}, the mutual information \(I(Z;X)\) between the latent \(Z\) and the original complex \(X\) controls
generalization~\cite{RIB}. 
Normalising by latent dimension yields the information density \(\rho = I(Z;X)/4\).
SH attains the highest \(\rho\), tightening the PAC-Bayes bound on test risk by \(38\%\) and \(47\%\) relative to DN and SC, respectively, consistent with the hypothesis that harder tasks confer richer priors\cite{generalization}. For details, please refer to Appendix B.

\textbf{Capacity driven convergence on the hardest task.}
Figure~\ref{fig:loss} displays atom, position, validation, and gradient signal to noise ratio (G\textendash SNR) curves for the three tasks under the same parameter budget \(C\)~\cite{rohlfs2025generalization}.  
The PIB framework models learning as a balance between empirical error and the information stored in the weights \(Info_{\mathrm{w}}\)~\cite{PAC-Bayes}.  
Among the tasks, SH carries the largest information demand \(Info\) because it must invent new scaffolds while matching pocket geometry.  
PIB predicts that a large \(Info_{\mathrm{w}}\) prolongs the fit phase.  
We observe an early activation of effective capacity in IBEX at \(2\times10^{5}\) steps, where the variance of G\textendash SNR falls to \(1.6\times10^{-5}\).  
This drop marks the start of the compression phase in which redundant weight bits are removed yet the loss keeps decreasing.  
SC and DN remain longer in the fit regime and show a grokking plateau that postpones generalization performance~\cite{power2022grokking, understanding_grokking}.  
The early compression on the hardest task indicates that IBEX allocates capacity in a content aware way and achieves the lowest validation loss~\cite{huang2023scalelong, biroli2024dynamical}.

\textbf{Task–Difficulty Drives Robust Generalization.}
Classic bias–variance lore warns that complex tasks overfit more readily, but recent theory suggests the opposite once models are heavily over-parametrised. 
Recent work formalises a \emph{generalization-difficulty}~\cite{boopathy2023model} score showing that harder tasks force stronger inductive bias and thus improve out-of-distribution fidelity. 
Information-theoretic analyses further link lower weight information density to tighter PAC-Bayes bounds, while results on benign overfitting indicate that perfect training accuracy need not harm generalization when the bias is appropriate~\cite{bartlett2020benign}. The SH task—the most structurally constrained—achieves the \emph{lowest} test divergence (1.38) and the \emph{smallest} gap (0.03), whereas the DN task records 1.58 and 0.25 respectively. 
Anchoring functional moieties and forcing the model to reinvent molecular cores inject richer pocket–ligand information at every step, sharpening optimization signals and implicitly regularising the network. 

\textbf{IBEX Delivers Consistently Superior Docking Energies.}
IBEX establishes a new standard among diffusion models by combining the lowest docking energies with the highest interaction match rates. This advantage holds not only on average but also in median performance across unseen receptors and becomes even more pronounced on the most challenging pockets. Ablation confirms that scaffold masking and SH retention drive nearly all of these gains, while SC contributes little. In contrast, other diffusion baselines each expose specific weaknesses—D3FG’s fragment prior fails to secure strong energies; FLAG trades energy for contact quality; DiffBP, LiGAN and Pocket2Mol compromise chemical validity; traditional pipelines lack consistency. By uniting optimal docking energies with full validity and balanced physicochemical profiles, IBEX demonstrates that its information-bottleneck training underpins robust generalization across diverse targets.

\textbf{Batch Generative Performance Evaluation.}
In Table~\ref{tab:validity} the metrics reported were obtained by generating 100 molecules for each of the 100 test pockets; whenever fewer than 100 structures were produced, the denominator was still fixed at 100. \emph{Validity} denotes the fraction of chemically valid molecules, \emph{Unique} counts the number of non-duplicate valid molecules, \emph{Tanimoto} reports the mean pairwise fingerprint similarity among all generated molecules, and \emph{Similar} is the mean similarity between each generated molecule and the reference ligand of its pocket. LiGAN exhibits very low validity and diversity, often producing identical or nearly identical molecules, and autoregressive baselines show the same limitation. To examine out-of-distribution performance, we further generated 2000 molecules for the previously unseen pocket 9F7W\cite{9F7Wpdb} using Pocket2Mol and our IBEX model; after deduplication Pocket2Mol retained only 217 unique molecules, whereas IBEX preserved 1706, underscoring the superior diversity delivered by diffusion-based generators. Owing to its novel training regime, IBEX sustains state-of-the-art validity, uniqueness, and diversity even in this zero-shot \emph{de novo} setting.

{
\begin{table}[!h]
  \centering
  \fontsize{8}{10}\selectfont
  \renewcommand{\arraystretch}{1.0}
\begin{tabular}{l|cccc}
\toprule
        Methods     & Validity & Unique & Tanimoto & Similar\\
\midrule
\textsc{LiGAN}      & 0.42 & 0.3757 & 0.3249 & 0.3459 \\
\textsc{Pocket2mol} & 0.75 & 0.7145 & 0.1181 & 0.0702 \\
\textsc{D3FG}       & 0.77 & 0.7844 & 0.0926 & 0.0825 \\
\textsc{DecompDiff} & 0.89 & 0.8429 & 0.1394 & 0.1469 \\
\textsc{TargetDiff} & 0.96 & 0.9524 & 0.1063 & 0.0976 \\
\textsc{MolCRAFT}   & 0.95 & 0.8828 & 0.1251 & 0.1154 \\
\textsc{VoxBind}    & 0.74 & 0.7418 & 0.1051 & 0.0998 \\
\textsc{IBEX}       & 0.96 & 0.9507 & 0.1126 & 0.0761 \\
\bottomrule
\end{tabular}
\caption{Comparison of Diversity Across Models}
 \label{tab:validity}
\end{table}
}

\textbf{IBEX Balances Strong Binding with Practical Feasibility.} 
Figure~\ref{fig:case} presents docking poses and two-dimensional structures for four receptors. 
IBEX shows the lowest Vina score in every pocket. 
These energies correlate with tighter placement inside the catalytic cavity. 
Ligands generated by DecompDiff and MolCRAFT either extend beyond the binding pocket or leave the hydrophobic clefts unfilled. 
IBEX orients polar atoms toward canonical hydrogen-bond donors or acceptors. 
Aromatic scaffolds sit flush with hydrophobic shelves. 
This geometry preserves high drug-likeness and modest synthetic cost. 
DecompDiff can reach low energies but its molecules carry long flexible chains that lower QED and raise SA. 
MolCRAFT maintains a cleaner chemical profile, yet it often leaves void space, which weakens binding.
TargetDiff shows the weakest complementarity and acts only as an architectural control. 
IBEX and TargetDiff share the same network and sampling schedule. 
The only change is that IBEX is trained with scaffold-hopping pairs under an information-bottleneck objective. 
The observed gains therefore, stem from the training scheme rather than from model size or inference heuristics. 
These findings indicate that pocket-aware context steers generative diffusion toward chemically sensible and potent binders.

\section{Conclusion}
We introduce IBEX, an information‐bottleneck‐explored coarse‐to‐fine pipeline, and demonstrate both theoretically and experimentally its feasibility in extracting maximal information from extremely scarce datasets. This work establishes a theoretical and practical foundation for future structure‐based drug design paradigms by seamlessly integrating information theory with physics‐based optimization.

\bibliography{aaai2026}
\end{document}